\documentclass{article}

\usepackage{PRIMEarxiv}

\usepackage[utf8]{inputenc} 
\usepackage[T1]{fontenc}    
\usepackage{hyperref}       
\usepackage{url}            
\usepackage{booktabs}       
\usepackage{amsfonts}       
\usepackage{nicefrac}       
\usepackage{microtype}      
\usepackage{lipsum}
\usepackage{fancyhdr}       
\usepackage{graphicx}       
\graphicspath{{media/}}     
\usepackage{changepage}
\usepackage{geometry}
\usepackage{amsmath}
\usepackage{amssymb}
\usepackage{siunitx}
\usepackage{multirow}
\usepackage{float}
\usepackage{lineno}
\usepackage{epsfig}
\usepackage{paralist}
\usepackage{comment}
\usepackage{subcaption}
\usepackage{orcidlink} 
\usepackage{tabularx}
\usepackage{csquotes}

\begin{document}
\pagestyle{fancy}
\thispagestyle{empty}
\rhead{ \textit{ }} 

\fancyhead[LO]{Enhancing Hyperspectral Image Prediction with Contrastive Learning in Low-Label Regimes}

\title{Enhancing Hyperspectral Image Prediction with Contrastive Learning in Low-Label Regimes}
\author{\Large
        Salma Haidar \textsuperscript{1,2,\thanks{Corresponding author: \texttt{salma.haidar@uantwerpen.be}}} \hspace{0.001cm} \orcidlink{0000-0003-4578-8877}
        \hspace{5cm} 
         Jos{\'e}  Oramas\textsuperscript{1} \orcidlink{0000-0002-8607-5067} \\
        \\
	\parbox{\textwidth}{\centering\textsuperscript{1} \textit{Department of Computer Science, University of Antwerp, imec - sqIRL/IDLab,\\ Sint-Pietersvliet 7,2000, Belgium}}\\
        \parbox{\textwidth}{\centering\textsuperscript{2} \textit{MicrotechniX BV,
        Anthonis de Jonghestraat 14 a, Sint Niklas,9100,Belgium}}\\     
 } 
 
\maketitle

\begin{abstract}
 Self-supervised contrastive learning is an effective approach for addressing the challenge of limited labelled data. This study builds upon the previously established two-stage patch-level, multi-label classification method for hyperspectral remote sensing imagery. We evaluate the method's performance for both the single-label and multi-label classification tasks, particularly under scenarios of limited training data. The methodology unfolds in two stages. Initially, we focus on training an encoder and a projection network using a contrastive learning approach. This step is crucial for enhancing the ability of the encoder to discern patterns within the unlabelled data. Next, we employ the pre-trained encoder to guide the training of two distinct predictors: one for multi-label and another for single-label classification. 
Empirical results on four public datasets show that the predictors trained with our method perform better than those trained under fully supervised techniques. Notably, the performance is maintained even when the amount of training data is reduced by $50\%$. This advantage is consistent across both tasks. 
The method's effectiveness comes from its streamlined architecture. This design allows for retraining the encoder along with the predictor. As a result, the encoder becomes more adaptable to the features identified by the classifier, improving the overall classification performance.
Qualitative analysis reveals the contrastive-learning-based encoder's capability to provide representations that allow separation among classes and identify location-based features despite not being explicitly trained for that. This observation indicates the method's potential in uncovering implicit spatial information within the data.
\end{abstract}


\section{Introduction}\label{sec:intro}

Hyperspectral image (HSI) analysis has garnered extensive attention over the past few decades, owing to its wide-ranging applications in precision agriculture, mineralogy, environmental science, medical diagnosis, cultural heritage, and defence~\cite{8314827}. 

\begin{figure*}[ht!]
\centering
\includegraphics[width=\textwidth]{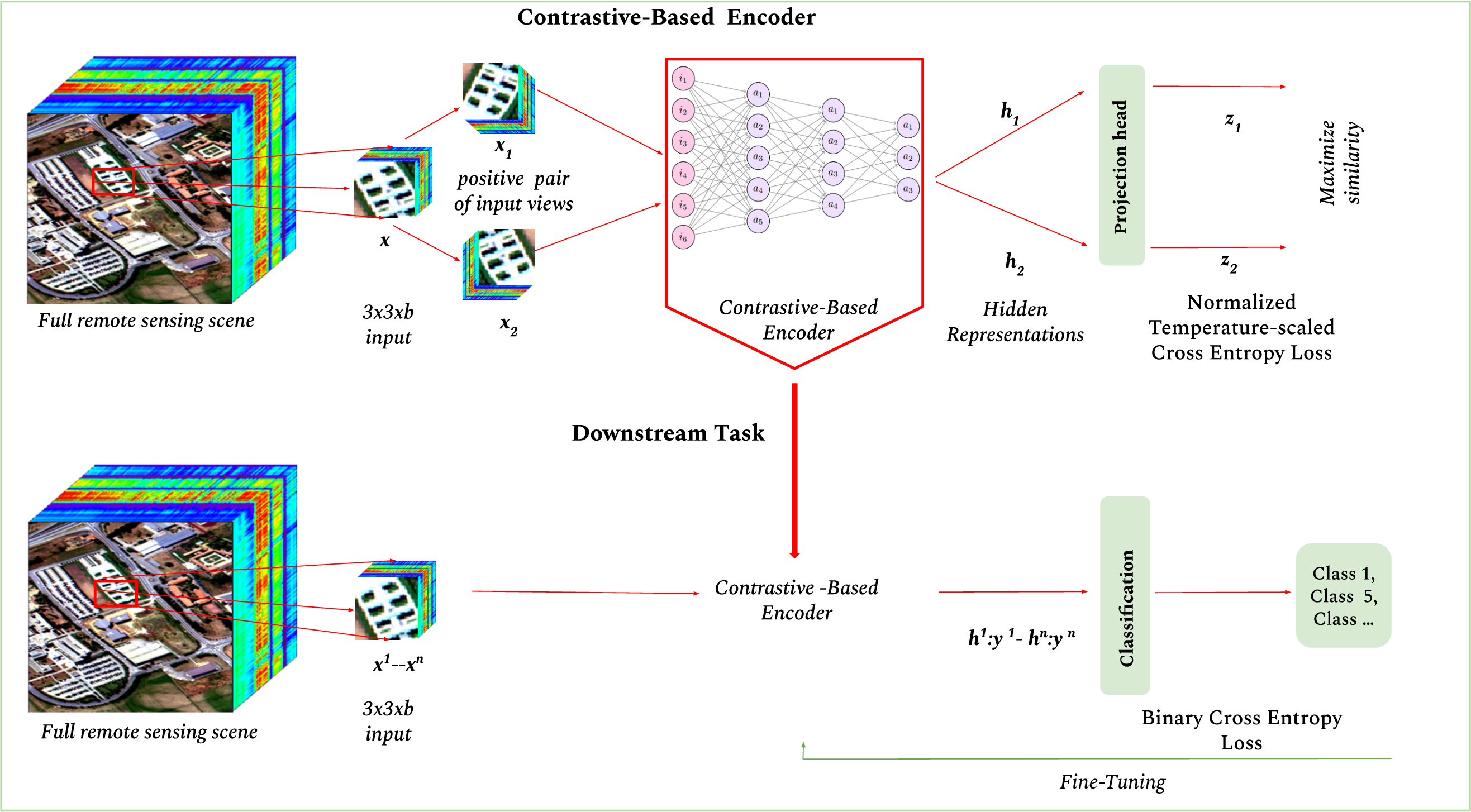} %
\caption{Two-stage method: In the first stage a contrastive-based encoder is trained using unlabelled data to maximise the agreement between hidden representations of two views of the same instance. In the second stage a classifier is fine-tuned with the pre-trained encoder on labelled examples for multi-label prediction.}\label{fig1:method}
\end{figure*}

The rapid advancements in imaging sensor capabilities have primarily driven the evolution of this field, significantly improving image quality by providing higher spatial and spectral resolutions. This technology captures detailed spectral signatures for each pixel through contiguous, narrow spectral bands. In hyperspectral remote sensing, this fine-grained spectral discrimination enables precise classification and analysis of land cover categories, such as forests, crops, and urban areas.

Due to the complexity and richness of hyperspectral data, deep learning algorithms are well-suited for processing and extracting relevant features from these intricate datasets. These algorithms can enhance the effectiveness of HSI in various applications, such as environmental monitoring, agriculture, urban planning, and mineral exploration~\cite{mou2017deep, yang2018hyperspectral, okwuashi2020deep}. Integrating deep learning with hyperspectral imaging expands the technology's potential across diverse disciplines.
However, fully leveraging hyperspectral data in deep models presents specific challenges. 

\noindent\textbf{Insufficient Labelled Data:} One of the primary challenges is the scarcity of labelled data required to train deep learning models effectively. Annotating hyperspectral images is often costly and labour-intensive due to the sheer volume of data, its inherent complexity, and the need for specialised knowledge.

\noindent\textbf{Architectural Sensitivity of Deep Networks}: The intricate sensitivity of deep learning networks to their architectures becomes more pronounced when dealing with hyperspectral data. The high dimensionality and richness of the data demand carefully designed neural networks that can effectively capture its complex spectral information.

\noindent\textbf{Complexities in Training Deep Models:} Training deep models on hyperspectral data presents significant challenges. Unlike traditional image data, hyperspectral images consist of numerous spectral bands, each contributing an extra layer of complexity. This high dimensionality increases the risk of overfitting, raises computational demands, and makes it harder for models to learn meaningful patterns. To address these issues, specialised training strategies—such as dimensionality reduction, advanced regularisation techniques, and custom loss functions—are crucial to enhancing the learning process and improving model performance.
 
Despite the advancements and strategies employed to tackle these challenges, multi-label classification is another aspect of hyperspectral image analysis that remains relatively underexplored.

Multi-label classification in hyperspectral image analysis remains under-explored compared to single-label classification, which dominates the literature. Typically, researchers divide large hyperspectral remote sensing scenes into smaller spatial patches and assign labels based solely on the label of the centre pixel. This methodology often neglects the contextual influence of neighbouring pixels, potentially diminishing the accuracy and completeness of the analysis.
Current research in this domain, such as~\cite{wang2016multi} and~\cite{10.1007/978-3-030-92238-2_19}, primarily focuses on pixel-level multi-label classification, which aligns more closely with unmixing algorithms. However, multi-label classification approaches are crucial for applications like land cover classification, where accurately predicting multiple objects within a scene can significantly enhance overall classification accuracy. This requirement is equally critical in microbiology, where bacteria frequently form confluent clusters, requiring methods capable of identifying multiple types within a single field of view.

The reliance of deep learning algorithms on extensive, accurately labelled datasets for robustness and generalisation poses a significant barrier in hyperspectral imaging applications. Manually curating these labels is labour-intensive and often falls short of providing the domain-specific accuracy required for reliable analysis. This challenge has led researchers to explore alternative learning paradigms, particularly within the domain of self-supervised learning (SSL)~\cite{9462394}, \cite{NEURIPS2019_a2b15837},~\cite{jing2020self}.
SSL is a method where the system learns to understand the data without the explicit guidance of labels. It distinguishes itself from unsupervised learning by generating its supervision through structured pretext tasks derived from the data itself. These tasks help the model learn the intrinsic characteristics of the unlabelled data. 
This approach enables the initial learning of general representations from the unlabelled data without being specifically tailored for any downstream task. However, these broadly learned representations can later be adapted or fine-tuned to align more closely with the specific downstream tasks of interest, such as classification, detection or other applications.
Among the various approaches within the SSL, Contrastive Learning (CL)~\cite{pmlr-v97-saunshi19a, 9226466} has emerged as an up-and-coming method. With its roots extending back several decades~\cite{carreira2005contrastive, 1640964}, the recent application of CL in self-supervised representation learning, especially in computer vision~\cite{OHRI2021107090, jaiswal2020survey}, has attracted significant attention. 
The essence of self-supervised contrastive learning lies in its ability to compare and distinguish data instances, enabling the generation of task-specific representations. This process generates pairs of augmented views of the same data sample, identifying common characteristics shared among similar instances while contrasting them with different views to emphasise the variations between distinct instances.
To accomplish this step, we apply various data augmentations to create multiple views of the same data instance, followed by employing contrastive loss functions~\cite{chen2021intriguing}. These functions effectively guide the model in distinguishing between similar and dissimilar instances, reinforcing the learning of meaningful and discriminative features. 
By leveraging this approach, Contrastive Learning facilitates learning general representations from unlabelled data, which researchers can fine-tune for the final intended task.

In the realm of hyperspectral data analysis, this methodology offers a compelling alternative to traditional supervised learning techniques, potentially overcoming the limitations posed by the high-dimensional and complex nature of the data.
In~\cite{10430726}, a deep yet streamlined encoder architecture for feature extraction and dimensionality reduction is introduced, paired with a classifier.
The approach leverages self-supervised contrastive learning for hyperspectral image analysis with the help of a simplified encoder design. Notably, it avoids complex deep learning structures such as ResNet, commonly used as a backbone architecture for feature extraction in contrastive learning frameworks. Despite its simplicity, the encoder from~\cite{10430726} outperforms existing state-of-the-art methods when fine-tuned with a classifier.
These results indicate that a straightforward architecture can achieve high performance in hyperspectral data processing.

Our work builds on the findings of~\cite{10430726} and explores the potential of contrastive learning for hyperspectral image analysis in scenarios with limited labelled training data, simulating real-world conditions. We fine-tune the pre-trained contrastive-based encoder alongside the classifier for multi-label and single-label classification tasks. This research makes the following additional contributions: 

\noindent\textbf{Robustness to Limited Data} We demonstrate the robustness of the contrastive learning-based method in extracting features from hyperspectral images as we gradually reduce the training data. Empirical results indicate that even with a $50\%$ reduction in data, the contrastive learning-based multi-label predictor maintains a competitive performance comparable to the supervised methods trained on the full dataset. 

\noindent\textbf{Single-Label Classification:} We utilise the contrastive learning-based architecture to perform single-label, patch-level classification on hyperspectral remote sensing scenes. Our results substantiate that the classifier fine-tuned with the contrastive learning-based encoder significantly outperforms traditionally supervised training methods on similar datasets without requiring any pre-processing techniques typically implemented in those methods.

\noindent\textbf{Qualitative Analysis of Hidden Representations:}
Our analysis reveals that the contrastive learning-based encoder captures contextual features and transitional properties more effectively than the supervised learning encoder within the jointly trained two-component network. Remarkably, this enhanced capability emerges without explicitly exposing the model to such information during training. As a result, the enhanced representation significantly improves the classifier's ability to distinguish among different classes of interest.
This insight offers a valuable understanding of the effectiveness of our approach, which is relevant not only for multi-label predictions but also for single-label predictions.

We organise the remainder of this paper as follows: Section~\ref{sec:relatedWork} reviews related work and positions our research with respect to existing efforts. Section~\ref{sec:methodology} describes the proposed method. In Section~\ref{sec:experiments}, we validate our method and conduct an empirical comparison with existing methods. In Section~\ref{sec:t-sne visualisation}, we conduct a qualitative analysis of the hidden representations generated by the contrastive-based encoder compared to those from supervised learning-based encoders. Finally, Section~\ref{sec:conclusion} offers concluding remarks and outlines prospects for future work.

\section{Related Work}
\label{sec:relatedWork}
Our work lies at the intersection of several key dimensions.
\subsection{Multi-label classification in hyperspectral image analysis}
Single-label, pixel-level classification is a well-explored topic in hyperspectral image analysis, largely due to the availability of public datasets with labelled pixels. Typically, researchers divide large hyperspectral remote sensing scenes into smaller patches and assign labels to each patch based on the centre pixel, neglecting the impact of the labels of neighbouring pixels. However, hyperspectral imaging sensors often have low spatial resolution, a trade-off for higher spectral resolution. This trade-off results in each pixel representing a larger area of the scene, often leading to mixed spectral signals when multiple materials are present within a single pixel. To address this trade-off, researchers have developed hyperspectral unmixing methods~\cite{6200362}, which decompose the measured pixel spectrum into a collection of constituent spectral signatures (endmembers) and corresponding fractional abundances. 
Multi-label classification in hyperspectral image analysis offers another solution to approach the mixed spectra issue, mainly when applied to patches extracted from larger images rather than individual pixels. Despite its potential, this approach remains under-researched. Existing studies have primarily focused on multi-label classification at the pixel level~\cite{10.1007/978-3-030-92238-2_19} and~\cite{wang2016multi}, which aligns more closely with unmixing algorithms.
Our work finds inspiration in the work presented in~\cite{rs15245656} and aims to leverage the advantages of patch-based multi-label classification within a self-supervised contrastive learning framework. By training the model on small patches of the hyperspectral image, each containing a subset of the scene, we seek to identify different materials present within a patch rather than within a single pixel. This patch-based approach includes the spatial context, potentially enhancing the classification accuracy by considering neighbouring pixels and mitigating the impact of mixed spectral signals at the individual pixel level. Additionally, patch-based multi-label classification can identify multiple materials within a small image region, providing a more comprehensive understanding of the scene. Analysing patches enables the model to capture spatial patterns and relationships between pixels, thereby enhancing the accuracy of material identification.
\subsection{Self-supervised contrastive learning (SSCL)} 
Self-supervised contrastive learning (SSCL) is a class of training methods that learn representations of data without labels. 
The main idea behind SSCL is to design a pretext task using positive and negative pairs of data samples. This approach trains the model to align the representations of augmented views of the same sample (positive pairs) while separating the representations of different samples (negative pairs)~\cite{pmlr-v139-wen21c}. This technique is applied in various domains, including image and video classification~\cite{Wu_2021_ICCV}, object detection~\cite{wu2022contrastive}, natural language processing \cite {zhang-etal-2022-contrastive-data}, among others.
SSCL shares the same metric learning objective as the Siamese networks~\cite{chen2021exploring} for learning representations that capture similarity or dissimilarity information.
Those networks are often used as the backbone architecture for contrastive learning tasks. \cite{e24040551} reviews Self-Supervised Learning (SSL) methods using auxiliary pretext and CL techniques. In~\cite{Wang_2021_CVPR}, a dense SSL method is proposed that targets the correspondence between local features (pixels). 
\cite {pmlr-v119-chen20j}, proposes a simple framework for contrastive learning of visual representations, \textit{SimCLR}, which is independent of the underlying architecture. Among the significant findings of this work is the pivotal role that robust data augmentation plays in contrastive prediction tasks. Moreover, using a non-linear transformation between the hidden representation and the contrastive loss improves the quality of the learned representations. 
Similarly, our research aims to uncover the hidden relationships within remote sensing hyperspectral data. We achieve this by designing a pretext task that enhances our understanding of the intricate interdependencies in the data. This task is structured to foster the generation of closely aligned representations for varying perspectives of the same data instance while ensuring that distinct data instances yield divergent representations.
Following this approach, the model learns general representations or embeddings without relying on explicit annotations. Instead, implicit labels are generated based on the inherent characteristics of the input data. In this context, we train an encoder to produce similar representations for the augmented views of the same hyperspectral image (HSI) patch sample.
\subsection{Self-supervised learning for Hyperspectral image analysis.}
The high dimensionality of the hyperspectral images, attributed to their extensive spectral information, poses challenges to traditional machine learning methodologies, such as Support Vector Machines (SVMs)~\cite{electronics12030488} and Random Forests, when utilised for hyperspectral image classification~\cite{MOUNTRAKIS2011247,RODRIGUEZGALIANO201293}. These approaches often require considerable feature engineering before applying learning algorithms, which can be cumbersome and inefficient. 

With the advent of deep learning methodologies, the field of hyperspectral image analysis witnessed substantial improvements.
Deep learning methods, particularly Convolutional Neural Networks (CNNs), have advanced hyperspectral image analysis by effectively handling the complex, high-dimensional data structures with greater precision~\cite{YUAN2020111716, tulapurkar2023multi, 10114986}. They excel at autonomously learning hierarchical feature representations from the data, reducing the reliance on manual feature engineering. This transition to deep learning has markedly enhanced the analytical capabilities of hyperspectral image analysis, achieving unprecedented accuracy and efficiency in image classification tasks.
Nonetheless, insufficient or scarce labelled training data remains a major hurdle. Even with advanced deep learning methods, this shortfall can lead to overfitting and restrict the generalisation capabilities of these models, thereby compromising the effectiveness of deep learning approaches in hyperspectral image analysis. This underscores the need for strategies to mitigate data limitations and improve model robustness.

One approach to address the challenge of limited labelled data in hyperspectral image classification is to develop deep learning models designed explicitly and specifically for this constraint~\cite{JIA2021179}. While "few-shot" classification can significantly curtail the time and effort required for data collection and labelling, such models remain prone to overfitting and limited generalisation. 
\cite{rs14215530} delves into using supervised contrastive learning (SCL) as a pre-training strategy for HSI classification. In this work, an encoder is pre-trained in a supervised setting using a combination of positive and negative samples, optimising the model parameters pairwise. However, the scarcity of labelled data continues to be a pressing issue. 
Alternatively, self-supervised learning offers a promising approach for hyperspectral image analysis by enabling effective representation learning independent of labelled data. For instance,~\cite{9734031} introduces a contrastive self-supervised learning (CSSL) algorithm that leverages Siamese networks. This method incorporates an augmentation module and a Siamese architecture to extract spatial-spectral features from hyperspectral images using minimal labelled data. Specifically, it utilises a ResNet-$18$ architecture and processes the data into spatial patches of $21\times21$. Labels are assigned to the centre pixel of each patch to fine-tune the classifier with the pre-trained ResNet network. 
Considering the abundance of unlabelled data,~\cite{hou2021hyperspectral} proposes a contrastive learning method for hyperspectral image (HSI) classification. The approach uses many unlabelled samples and employs data augmentation techniques. Furthermore, ~\cite{CAO202171} proposes an unsupervised feature learning method based on autoencoders and contrastive learning. The method aims to extract better features for the classification of pixel-level, single-label hyperspectral images. 
In response to the challenge of limited labelled samples in the hyperspectral imaging data, ~\cite{doi:10.34133/remotesensing.0025} proposes a Deep Contrastive Learning Network (DCLN). First, Principal Component Analysis (PCA) is applied to reduce the spectral dimensionality of the data. Next, pixel cubes of size $11\times11\time20$ are extracted to form patches.
The similarity between the centre pixel and the neighbouring pixels in each patch is then calculated, with only the top \textit{m} pixels exhibiting the highest similarity considered for further processing while the rest are set to zero. Subsequently, two contrastive groups are formed using the labelled data samples to train the feature extraction network through contrastive learning techniques. 
Unlike methods that generate two views of the same data sample, this method groups samples with the same label into distinct contrastive groups while preserving their positional context. Once trained, the network extracts spectral-spatial features from hyperspectral image pixels and generates pseudo-labels for each unlabelled sample based on their spatial-spectral mixing distance. The network then incorporates samples with high-confidence pseudo-labels into the training set, allowing fro retraining and improved performance. 
Similarly,~\cite{9664575} addresses the issue of limited labelled samples in hyperspectral imagery by adopting a self-supervised learning paradigm. Their method introduces a two-stage training approach, combining contrastive learning with supervised fine-tuning to enable classification with minimal labelled data. During pre-processing, the hyperspectral images are divided into smaller blocks $23\times 23$. First, they perform dimensionality reduction using Principal Component Analysis (PCA). Next, they apply an Extended Morphological Profile (EMP) to capture spatial features. Additionally, they employ data augmentation techniques to further improve model performance. 
~\cite{lee2022self} proposes an architecture that leverages cross-domain convolutional neural networks. This architecture incorporates shared parameters to learn representations across hyperspectral datasets with varying spectral characteristics and without pixel-level annotations.
In~\cite{rs15123123}, the authors introduce optimisation strategies to enhance feature extraction and classification performance within the contrastive learning framework. These strategies encompass various data augmentation techniques, such as band removal, random occlusion, and gradient masking. Additionally, they emphasise the preservation of spatial-spectral information by adopting patch-based data instances. In this context, each patch is assigned the label of its centre pixel. This approach accentuates the importance of spatial-spectral details from neighbouring pixels adjacent to the centre while concurrently diminishing the influence of more distant pixels.

Our research employs self-supervised contrastive learning techniques for hyperspectral image analysis, aligning with previous studies. However, we make several unique contributions. First, we address the patch-level multi-label prediction task, an area that has received limited attention in the literature, and evaluate its effectiveness under scenarios with limited data availability. This allows for the simultaneous identification of multiple materials within each patch, enhancing classification comprehensiveness. Secondly, we adapt our approach to single-label classification tasks and assess its performance in data-constrained regimes, demonstrating the method's versatility. Finally, we compare our single-label method with existing patch-level, single-label classification methodologies, highlighting improvements in accuracy and robustness. Like these methodologies, our approach emphasises patches with labels corresponding to the centre pixel. These contributions collectively enhance the robustness and versatility of hyperspectral image classification, particularly in data-constrained settings.

\section{Methodology}
\label{sec:methodology}
Figure~\ref{fig1:method} provides a schematic representation of the proposed methodology, comprised of a sequential two-stage process. We employ a contrastive learning technique to train a base encoder in the initial stage. This step ensures that similar patches will yield analogous representations, thereby preserving the inherent structure of the data.
This stage comprises a tripartite process: firstly, we create two augmented views for each patch. Secondly, these views are mapped to a latent space using the trained encoder, resulting in two intermediate representations.
Furthermore, we use a neural network with two fully connected layers to map the intermediate representations into vector embeddings. The projection network, comprising these fully connected layers, then maps these vectors into a space where a contrastive loss function is applied to maximise the distinction between them and enhance the effectiveness of the learning process. This setup ensures that similar samples are closely clustered in the embedding space while dissimilar samples are well-separated, improving the model's ability to generalise from the learned representations.
In the next stage, after discarding the projection network and combining the pre-trained encoder with a classifier, we fine-tune the entire network using the labelled examples in two distinct scenarios, as detailed in Section~\ref{subsec:classifier-performance}. 
The encoder will generate low-dimensional hidden representations that capture the essential features of the input data. Concurrently, we train the classifier to discern discriminative features within these representations for multi-label prediction. 

\subsection{Model Architecture}
\label{subsec:model_design}
\textbf{Contrastive Learning Feature Extraction Network.} It comprises several components.
The augmentation component applies random vertical and horizontal transformations to an input sample, in our case, a patch, creating two views referred to as a positive pair. Each view passes through a network consisting of fully connected layers with Rectified Linear Unit (ReLU) activation and dropout layers. This network component serves as the encoder, responsible for extracting informative features from the two augmented views of the same input sample and mapping them into two hidden (intermediate) representations $h_{1}$ and $h_{2}$.
The mapping function is defined as $h_{i} {=} f(W_h\cdot x_{i} + b_h )$ where $x_{i}$ represents an augmented view of the input samples $X^{M}$ (${M}$ ${=}$ the total number of samples), and $W_h$ and $b_h$ being the weights and the bias of the encoder, respectively. The encoder will preserve the spatial dimension of the samples yet reduce the spectral dimension. 
The projection head is composed of two fully connected layers with ReLU and dropout layers. Its purpose is to project the hidden representations, $h_{1}$ and $h_{2}$ , into another space represented by $z_{1}$ and $z_{2}$.
The mapping function is defined as $z_{i} = g(W_z\cdot h_{i} + b_z )$ where $z_{i}$  denotes the vector representation of the intermediate representation, and $W_z$ and $b_z$ represent the weights and the bias of the projection network, respectively. 
During training, the contrastive loss function will direct the weights to update towards maximising the similarity between the two vector representations {$(z)$}. As a result, the encoder produces two similar hidden representations {$(h)$} for the augmented views generated from the same sample, facilitating the model's ability to recognise and differentiate between nuanced features within the data.
Ultimately, this will enable it to learn the relevant features present in the input samples. 
In this context, we employ the Normalised Temperature-scaled Cross Entropy Loss (NT-Xent)~\cite {10.5555/3157096.3157304, https://doi.org/10.48550/arxiv.2205.03169}, as our contrastive loss function.
The objective of NT-Xent (Eq~\ref{eq: CLLOSS}) is to increase the similarity between the two augmented views (positive pair) while simultaneously asserting their dissimilarity with the negative samples. The negative samples comprise the remaining augmented views generated from the other input samples or patches within the batch. Let $N$ represent the batch size; given a positive pair, the remaining $2(N{-}1)$ views serve as negative pairs. According to the approach in~\cite{pmlr-v119-chen20j}, this ensures proper pairing during the process.
Prior to starting the training process for the contrastive learning model, we carefully examined the patches in our dataset, removing the duplicates that appeared within each batch.
 
\begin{equation} 
\begin{aligned}
&l_{i,j} = -\log \left( \frac{\exp\left(\frac{\text{sim}(z_i, z_j)}{T}\right)}{\sum_{k=1}^{2N} \mathbb{1}_{[k \neq i]} \exp\left(\frac{\text{sim}(z_i, z_k)}{T}\right)} \right)
\end{aligned}
\label{eq: CLLOSS}
\end{equation}

In Eq~\ref{eq: CLLOSS}, $sim(z_i,z_j)$ is the cosine similarity $\frac{z_i^{T}z_j}{\|{z_i}\|\|{z_j}\|}$, and since contrastive learning involves two views of each patch, there are 2$N$ data points. The function $\mathbb{1}_{[k\not=i]}\in {\{0,1\}}$ evaluates to $1$ if $k$ is not equal to $i$ and $0$ otherwise. The temperature scale $T$ is a constant used to scale the cosine similarity values, ensuring they are not too large or too small. 
The similarity between the augmented samples is calculated pairwise for $(i,j)$ projections, encompassing all $2N$ projections within the entire batch. The overall loss function, used for back-propagation, is the average taken across all positive pairs in the batch shown in Eq~\ref{eq: CLLOSS_for_batch} :
\begin{equation} 
\begin{aligned}
&Loss = \frac{1}{2N}\sum_{k=1}^{N}[l(2k-1, 2k) + l(2k, 2k-1)]
\end{aligned}
\label{eq: CLLOSS_for_batch}
\end{equation}

\textbf{Downstream Network}.
After training the base encoder with a contrastive learning approach, we integrate it with a classifier to adapt it for the classification task. We employ the same classifier architecture used in~\cite{rs15245656}, which consists of fully connected layers, non-linear activation function, and dropout layers. It is designed for patch-level classification of hyperspectral data.
The base encoder generates a hidden representation {$h_{i}$} for each input sample. The classifier receives these representations and predicts the associated class(es). To optimise the multi-label classification model, we use Binary Cross Entropy with Logits Loss (Eq.~\ref{eq: BCELoss}), a variant of the binary cross-entropy loss applied independently to each class.
\begin{equation} 
\begin{aligned}
 &\ell_c(\hat{y},y) = \{l_{1,c},\dots,l_{N,c}\}^\top\\
\end{aligned}\label{eq: BCELoss}\end{equation}

Where $\ell_c(\hat{y}, y)$, aggregates the individual loss values, computed using Eq.~\ref{eq: BCELosspersample} across all instances $N$ of class $c$, into a single vector. $\top$ is the transpose operator, which organises the set of losses into a column vector.
 
\begin{equation} 
\begin{aligned}
&l_{n,c} = -w_{n,c}[p_c y_{n,c}\cdot\log\sigma(\hat{y}_{n,c})+(1 - y_{n,c}) \cdot\\&\log(1 -\sigma(\hat{y}_{n,c}))]\end{aligned}\label{eq: BCELosspersample}\end{equation}
\ 
$l_{n,c}$ represents the loss for the \(n\text{-th}\) sample in the batch with respect to the \(c\text{-th}\) class. 
In this context, $n$ denotes the sample index within the batch, and $c$ denotes the class index.
$w_{n,c}$ represents the weight for ${n}$ and ${c}$. $p_c$ is the weight of the positive outcome for class $c$. It adjusts the impact of positive examples in the loss function, often used to handle class imbalance by giving more importance to underrepresented classes.
$y_{n,c}$ is the ground truth label for the example $n$ and class $c$, it takes value from the label space $y=\{0,1\}^c$, where $0$ indicates the absence and $1$ indicates the presence of the class.
$\hat{y}_{n,c}$ is the predicted score (logit) for example $n$ belonging to class $c$. It is the output of the model before applying the sigmoid function $\sigma(.)$, (\(\sigma(x) = \frac{1}{1 + e^{-x}}\)).

For the single-label classification task, we implement the Cross-Entropy Loss defined in Eq.~\ref{eq: CELoss}
\begin{equation}
\begin{aligned}
L = -\frac{1}{N} \sum_{i=1}^{N} \sum_{j=1}^{C} y_{ij} \log(p_{ij})
\end{aligned}\label{eq: CELoss}\end{equation}

$L$ represents the total loss across all samples. $N$ is the number of samples in the dataset, and $C$ is the number of the classes. $y_{i,j}$ represents a binary indicator if the label $j$ is the correct classification for sample $i$. $p_{i,j}$ is the predicted probability that sample $i$ belongs to class $j$. The outer sum iterates over all samples, and the inner sum iterates over all classes for each sample. Finally, the logarithmic function penalises the predicted probabilities based on their distance from the true label.

\section{Experiments}
\label{sec:experiments}

\begin{figure}[htbp]
    \centering
    \begin{subfigure}[t]{0.45\textwidth}
        \centering
        \includegraphics[width=\textwidth]{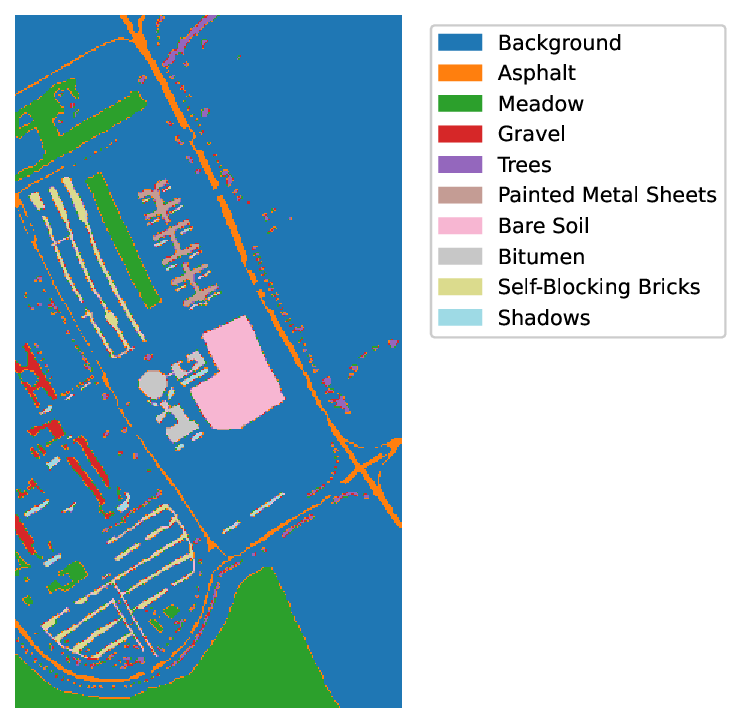}
        \caption{Pavia University Ground Truth}
        \label{fig:a-paviau}
    \end{subfigure}
    \hfill
    \begin{subfigure}[t]{0.45\textwidth}
        \centering
        \includegraphics[width=\textwidth]{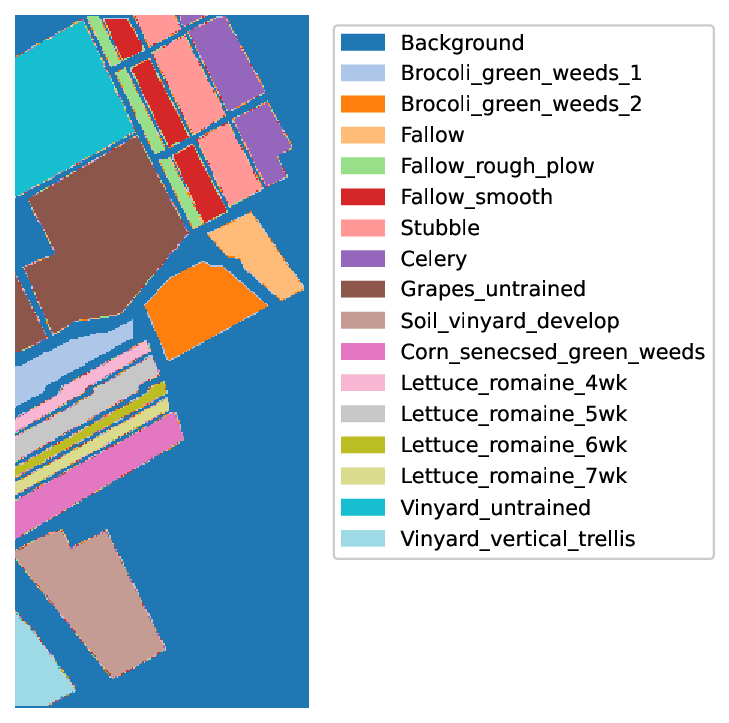}
        \caption{Salinas Ground Truth}
        \label{fig:b-salinas}
    \end{subfigure}
    \caption{Pavia University and Salinas ground truth visualisations.}
    \label{fig2: paviau_salinas_visualisation}
\end{figure}

\begin{figure}[htbp]
    \centering
    
    \begin{subfigure}[t]{0.45\textwidth}
        \centering
        \includegraphics[height=3cm,width=\textwidth]{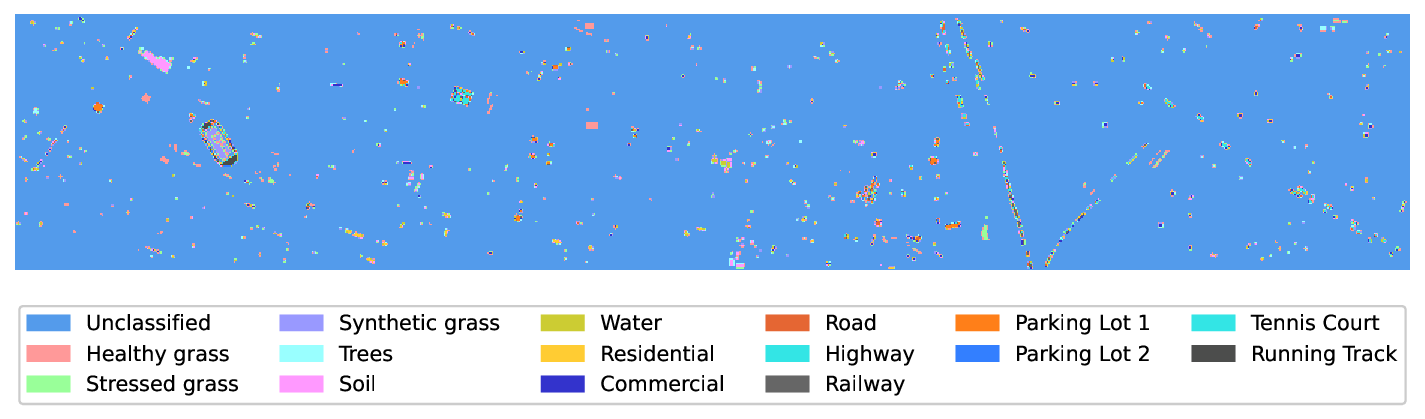}
        \caption{Houston 2013 Ground Truth}
        \label{fig:a-houston2013}
    \end{subfigure}
    \hfill
    \begin{subfigure}[t]{0.45\textwidth}
        \centering
        \includegraphics[width=\textwidth]{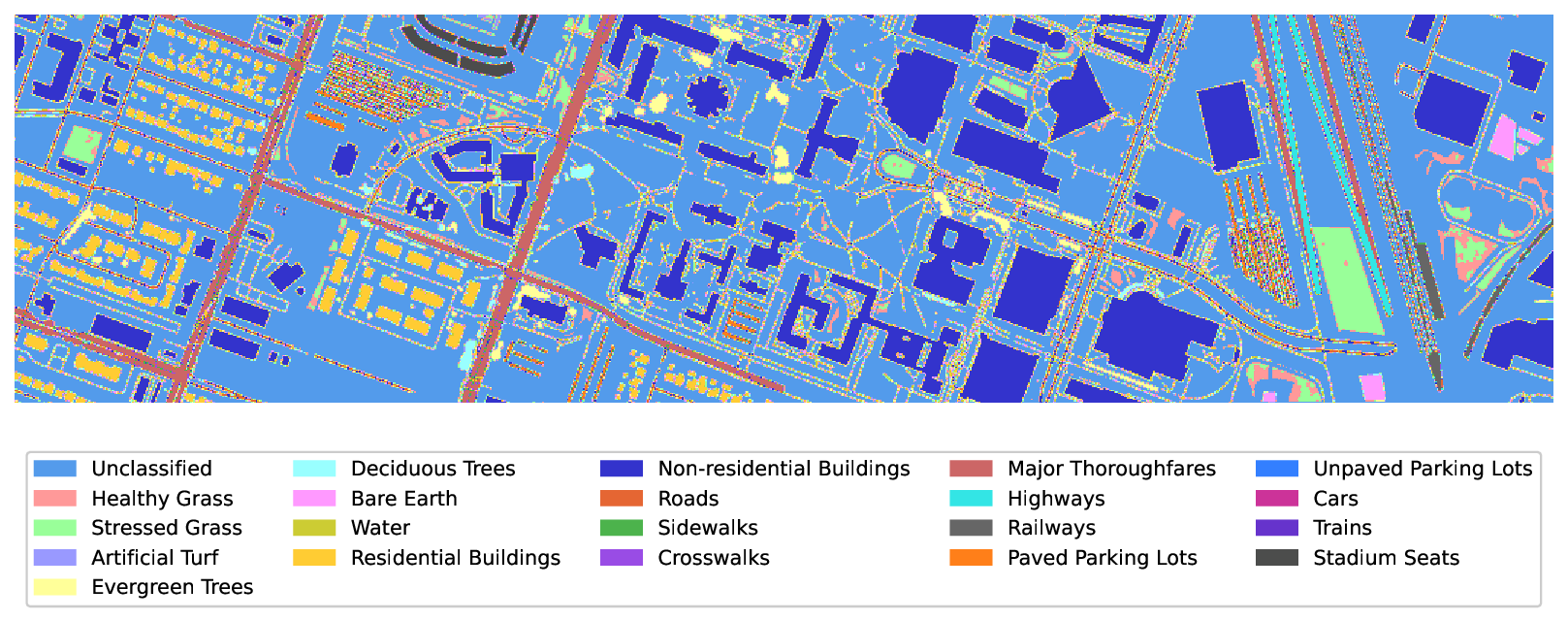}
        \caption{Houston 2018 Ground Truth}
        \label{fig:b-houston2018}
    \end{subfigure}
    \caption{Houston 2013 and Houston 2018 ground truth visualisations.}
    \label{fig3:Hosutons_visualisation}
\end{figure}

In this section, we report the outcomes of our experiments, utilising four publicly available datasets, Pavia University (PaviaU), Salinas~\cite{PaviaUSalinas}, Houston 2013~\cite{Houston2013} and Houston 2018~\cite{Houston2018}. Figures~\ref{fig2: paviau_salinas_visualisation} and~\ref{fig3:Hosutons_visualisation} present a pseudo-colour ground truth overview of the four datasets. All four datasets follow a consistent structure in the form of $X\in\mathbb{R}^{h,w,b}$, where $h, w$, and $b$ represent the height, width, and number of spectral bands, respectively. It is essential to highlight that these datasets are widely used in hyperspectral image analysis and are representative of the scale typically available in public datasets within the field of remote sensing ~\cite{9007624, 9440852, 9325094,9645266}. Their widespread use reflects both their prevalence and the standard scale for research in this domain.
\vspace{0.5cm}

\noindent\textbf{Pavia University (PaviaU)}\\
 The PaviaU dataset, Figure~\ref{fig:a-paviau}, was collected by the Reflective Optics System Imaging spectrometer (ROSIS-3) sensor over Pavia, Italy. It spans a spectral range of $430nm-860 nm$ with a spatial resolution of $1.3$ meters per pixel. The scene has the dimensions of $(610{\times}340{\times}103)$, and it contains a total number of $42,776$ labelled pixels. It comprises $9$ classes along with a background class. The ground truth includes asphalt, meadows, gravel, trees, metal sheets, bare soil, bitumen, brick, and shadow.\\
 
\noindent\textbf{Salinas}\\
 The Salinas dataset was captured by the 224-band AVIRIS sensor over Salinas Valley, California (Figure~\ref{fig:b-salinas}). The spectral coverage ranges from $400 nm$ to $2500 nm$, and the dataset features a high spatial resolution of $3.7 -$ meter pixels. After disregarding $20$ water absorption bands, the final dataset dimensions are $(512{\times}217{\times}204)$. It contains $16$ classes and a background class. Ground truth covers crops, vegetables, bare soil, and vineyard fields with a total of $54,129$ labelled pixels.\\

\noindent\textbf{Houston 2013}\\
The Houston 2013 dataset, Figure~\ref{fig:a-houston2013}, was provided as part of the 2013 IEEE GRSS Data Fusion Contest – Fusion of Hyperspectral and LiDAR Data. It includes a hyperspectral image with a spectral range of $380nm-1050 nm$ and a spatial resolution of $2.5$-meters per pixel. The dimensions of the scene are $(349{\times}1905{\times}144)$. The dataset was captured over the University of Houston campus and the neighbouring urban area containing 15 classes and an unclassified background class. Ground truth covers various urban features including healthy, stressed, synthetic grass, trees, soil, water, residential and commercial buildings, roads, highways, railways, parking lots, tennis courts and, running tracks.\\

\noindent\textbf{Houston 2018}\\
The Houston 2018 dataset, Figure~\ref{fig:b-houston2018}, was introduced as part of the 2018 IEEE GRSS Data Fusion Challenge – Fusion of Multispectral LiDAR and Hyperspectral Data. Like Huston 2013, it was collected over the University of Houston campus and its surrounding area, with the same spectral coverage of $380nm-1050 nm$. However, the Houston 2018 dataset includes 48 bands, compared to $144$ in Houston 2013. 
The dataset dimensions are ($1202 {\times} 4172 {\times} 48$), and the ground truth includes $20$ classes along with an unclassified background class. Of these classes, $15$ overlap with those in Houston 2013, while the remaining five provide more detailed land cover information. 

Houston 2018 includes $2,021,190$ labelled pixels.  Processing this large dataset would generate nearly $200,000$ patches of size $3{\times} 3{\times} b$, leading to significant computational demands. To address this, we utilised a subset of the dataset as provided by~\cite{9540028}. This subset focuses on an overlapping region between Houston 2013 and Houston 2018 to ensure consistency. Given that Houston 2018 has a higher spatial resolution ($1$ meter per pixel) compared to Houston 2013 ($2.5 $ meter per pixel), the area covered by $210 {\times} 954$ pixels in Houston 2018 corresponds to a smaller region in Houston 2013 but offers greater detail and quality. This approach aligns with our objective to demonstrate the efficiency of our method on smaller datasets while leveraging the higher-quality hyperspectral data provided by Houston 2018. The subset we utilised has the following $7$ classes: residential and commercial buildings, trees, water, healthy and stressed grass, and roads.\\

From these datasets, we extract patches of size $p\in\mathbb{R}^{h'{\times} w'{\times}{b}}$ following the sampling paradigm from~\cite{rs15245656}, thereby reducing the spatial dimension while preserving the number of spectral bands. The resulting patch dimensions are $(3,3,b)$. Notably, this patch extraction procedure does not allow overlapping, resulting in a significant reduction in the number of patch data samples compared to the number of labelled pixels in the original scenes. 
We selected the $3\times 3$ spatial size for our patches based on the characteristics of the datasets we are using and the fact that we are imposing a stride to prevent overlapping while maintaining a count of data instances that can be reasonable to train our method.
Our experiments predominantly investigate the impact of contrastive learning on two tasks: multi-label and single-label patch-level hyperspectral classification. In this regard, we assign labels to these patches under both scenarios. For multi-label, assigned annotations indicate the presence of mixed classes within the patch. We excluded patches uniformly composed of the background class. However, the patch is included in the dataset if the background class appears alongside other classes. Following the common practice, we assign a label corresponding to the class of the centre pixel for the single-label classification task. Patches with centre pixels representing the background class, or those entirely composed of the background class, are ignored. 

Table~\ref{tab:patches_distribution} summarises the distribution of patch types generated through the multi-label and single-label sampling processes. 
In the multi-label patch datasets, a substantial proportion of the patches exhibit significant class diversity. Specifically, approximately $55\%$, $82.85\%$ and $63.5\%$ of the patches from PaviaU, Houston 2013 and Houston 2018, respectively, contain mixed class labels, indicating heterogeneity and diversity of patterns in the data. In contrast, the Salinas dataset shows less class mixing, with only $21\%$ of patches containing multiple classes. This comparison underscores the distinctive characteristics of each dataset and their potential impact on classification performance.

In the single-label sampling process, most patches are uniform, containing only one class. However, some patches may include mixed-class pixels, though the class label is determined by the central pixel. These mixed patches are less prevalent than in the multi-label scenario, as patches predominantly composed of background pixels are typically excluded from the analysis. 
 
\begin{table*}[htbp]
\centering
\caption{Patches sampled from the datasets using multi-label and single-label sampling procedures.}
\label{tab:patches_distribution}
\setlength\tabcolsep{8pt}  
 
\begin{tabular*}{\textwidth}{@{\extracolsep\fill}lccccccccc}
\toprule
{} && 
{\textbf{\textit{PaviaU}}} &
{\textbf{$\%$}} &
{\textbf{\textit{Salinas}}} &
{\textbf{$\%$}} &
{\textbf{\textit{Houston 2013}}} &
{\textbf{$\%$}} &
{\textbf{\textit{Houston 2018}}} &
{\textbf{$\%$}}\\
\midrule
\multicolumn{10}{l}{\textbf{\textit{Multi-Label Sampling}}} \\
\textit{Mixed patches} && 3774 & 55 & 1442 & 21 & 2710 & 82.8 & 6341 & 65 \\
\textit{Uniform patches} && 3125 & 45 & 5289 & 79 & 561 & 17.2 & 3422 & 35 \\
\textbf{\textit{Total}} && 6889 &  & 6731 &  & 1665 &   & 9763 &   \\
\hline
\multicolumn{10}{l}{\textbf{\textit{Single-Label Sampling}}} \\
\textit{Mixed patches} && 1742 & 36 & 721 & 12 & 1104 & 66.4 & 2638 & 44.6 \\
\textit{Uniform patches} && 3097 & 64 & 5290 & 88 & 561 & 33.6 & 3282 & 55.4 \\
\textbf{\textit{Total}} && 4839 &   & 6011 &   & 1665 &  & 5920 &   \\
\hline
\end{tabular*}
\end{table*}
 
We designed a series of experiments to evaluate the performance of our classifier under the constraints of limited datasets. Two distinct experimental sets were conducted: one focusing on multi-label classification and the other on single-label classification. This analysis is crucial for assessing the robustness and adaptability of our classifier, which is especially important in real-world applications where data availability is often limited or constrained.
In addition, we conducted a series of tests to evaluate the model's performance under various conditions. Specifically, we assessed the classifier's performance with different temperature parameters in the contrastive learning loss function. We also investigated the impact of varying patch sizes on the classifier's accuracy, comparing it to other methods that utilise the same patch sizes. Finally, we explored the effect of different hidden representation sizes in the contrastive learning encoder to understand their influence on model performance.

\subsection{Implementation Details}
In our model's self-supervised contrastive learning (SSCL) phase, we generated two views for each input sample by applying horizontal and vertical flips. We then subjected these transformed samples to z-score normalisation.

In the initial phase of our model training, specifically for the base encoder, we split our randomly shuffled patch dataset into two portions: $90\%$ designated for training and $10\%$ for validation. We adopted a different data partitioning approach for the subsequent training phase centred on the classifier. After subjecting the entire dataset to a random shuffle, approximately $10\%$ was reserved as a test set. The remaining data was then randomly split, allocating $80\%$ to training and $10\%$ to validation.

Throughout our experiments, we consistently employed the Adam optimisation~\cite{DBLP:journals/corr/KingmaB14} for updating the network weights during training. Additionally, we utilised the StepLR learning rate scheduling method, applying a $\gamma$ reduction after a specific number of epochs. Each hyperparameter — including the learning rate schedule, batch size, and number of epochs — was meticulously optimised through extensive experimentation to enhance the performance of each baseline. For a comprehensive overview of the various hyperparameters employed in training the contrastive learning-based encoder and the classification models for both multi-label and single-label classification tasks, please refer to the Supplementary Material document (Section $2$). To ensure reproducibility and minimise the effect of randomness, we averaged the accuracy reported in all our experiments over three independent runs, each initiated with a distinct random seed.

\subsection{Multi-Label Classification}
\label{subsec:multi-label-classification}

In this experiment, we conducted patch-level, multi-label classification on hyperspectral images. For that purpose, we trained the classifier alongside the contrastive learning-based encoder and evaluated the performance under two scenarios: 

\noindent\textit{\textbf{CL-tune}}: In this scenario, the layers of the base encoder were unfrozen, allowing encoder's weights to be retrained during the classification task.

\noindent\textit{\textbf{CL-freeze}}: In this scenario, the layers of the base encoder were frozen, meaning the encoder's weights were fixed and not updated during training.  

We report performance using average accuracy~\cite{sorower2010literature}, and compare our results with those obtained from three training schemes outlined in~\cite{rs15245656}.

\noindent\textit{\textbf{Iterative}} scheme where the autoencoder and classifier function as separate architectures with distinct objectives. It bears a resemblance to the iterative training employed in generative adversarial networks (GANs).

\noindent\textit{\textbf{Joint}} scheme~\cite{LIU2022107007}, the autoencoder and classifier integrated into a single, unified model trained end-to-end.

\noindent\textit{\textbf{Cascade}} scheme~\cite{xing2016stacked}, the autoencoder is first pre-trained independently to reconstruct the input data; subsequently, the encoder is then used together with a classifier, with the encoder's layers frozen, meaning no fine-tuning occurs.

All these schemes share a comparable encoder architecture, featuring a hidden layer with $32$ neurons. The classifier architecture is also similar across the schemes, but the output layer is adjusted to accommodate the varying number of classes in each dataset.
We did not directly compare our method with existing contrastive learning approaches for HSI in the literature, given their engineered design for a pixel-level, single-label classification task. Adapting these methods for our patch-level, multi-label classification task would require extensive modifications, which would compromise the integrity of the original methods.

\subsubsection{Performance Across Different Training Schemes}
\label{subsec:classifier-performance}
\begin{table*}[htbp]
\centering
\caption{Accuracy performance ($\%$) of the CL-based multi-label classifier compared to three fully supervised learning schemes~\cite{10430726}.}\label{tab:model_acc_h32}%
\setlength\tabcolsep{8pt}

\begin{tabular*}{\textwidth}{@{\extracolsep\fill}lcccccc}
\toprule
& \textbf{\textit{Iterative}} & \textbf{\textit{Joint}} & \textbf{\textit{Cascade}} & \textbf{\textit{CL-freeze}} & \textbf{\textit{CL-tune}} \\
\midrule
\textbf{\textit{PaviaU}}&$84.03 $&$86.14 $&$83.50 $&$70.56$&$\textbf{87.87}$\\
\textbf{\textit{Salinas}}&$87.61 $&$86.40 $&$86.47 $&$74.90$&$\textbf{88.86 }$\\
\textbf{\textit{ Houston 2013}}&$85.81 $&$ 85.92$&$56.12$&$59.84$&$\textbf{87.23}$\\
\textbf{\textit{Houston 2018}}&$85.59 $&$\textbf{86.03}$&$60.21$&$59.27$&$85.28$\\
\hline
\end{tabular*}

\end{table*}

In Table~\ref{tab:model_acc_h32}, our method demonstrates superior performance compared to other schemes when we retrain the contrastive learning-based encoder alongside the classifier (\textit{CL-tune}). 
On the PaviaU dataset, performance improvements ranged from $1.73\%$ to $4.37\%$. For Salinas, these improvements ranged from $1.25\%$ to $2.46\%$. Considering the complexity and diversity of mixed patches in the PaviaU dataset, we infer that contrastive learning is particularly well-suited for modelling such data, especially considering the dataset's relatively small size. 

A similarly noteworthy pattern emerges when comparing performance on the Houston 2013 and Houston 2018 datasets. Although the Houston 2013 dataset is significantly smaller, our method outperformed its results on Houston 2018 and showed considerable improvements over supervised learning schemes. Houston 2013 contains a higher proportion of mixed patches compared to uniform patches (Table~\ref{tab:patches_distribution}), indicating greater variability within each patch. The strong performance of our method on the Houston 2013 dataset further supports the conclusion drawn from the PaviaU results, as both datasets share similar characteristics.

Notably, the \textit{CL-tune} variant outperforms the \textit{CL-freeze} variant by a significant margin of $17.31\%$ and $13.98\%$ on the PaviaU and Salinas datasets, respectively. However, when evaluating the results on the Houston datasets, we observe a significant degradation in performance, similar to the results of the Cascade method. The Cascade and the \textit{Cl-freeze} share the structural feature of pre-training the encoder separately and freezing its layers during integration with the classifier. While this approach reduces training time, it restricts the model's ability to adapt to new data and capture finer details. Consequently, performance tends to degrade, particularly on more complex datasets like Houston and PaviaU.
The improved performance on the \textit{CL-tune} suggests that retraining the encoder allows the weights to update more effectively in alignment with the features learned by the classifier. These features capture the relevant information necessary for accurate label prediction. The prediction error is back-propagated to the earlier layers, namely those of the encoder,  enabling continuous refinement. This retraining approach offers the advantage of using the pre-trained weights of the base encoder as a robust initialization, which facilitates more efficient learning and improves the representations of the encoder. 

\subsubsection{Impact of the Amount of Labelled Data}
\label{subsec: ml-amount-of-data}

In Section~\ref{subsec:classifier-performance}, we confirmed that the classifier's performance was higher when operating under the contrastive learning paradigm than the supervised learning paradigm. This disparity raised an intriguing question: What would happen if we reduced the amount of labelled data used for fine-tuning the classifier with the contrastive learning-based encoder? How would such a reduction impact the overall performance?

To address this question, we conducted an experiment to assess the impact of gradually decreasing the available labelled training data on the performance of the classifier.

\begin{figure}[htbp]
\centering
\includegraphics[width=0.8\linewidth]{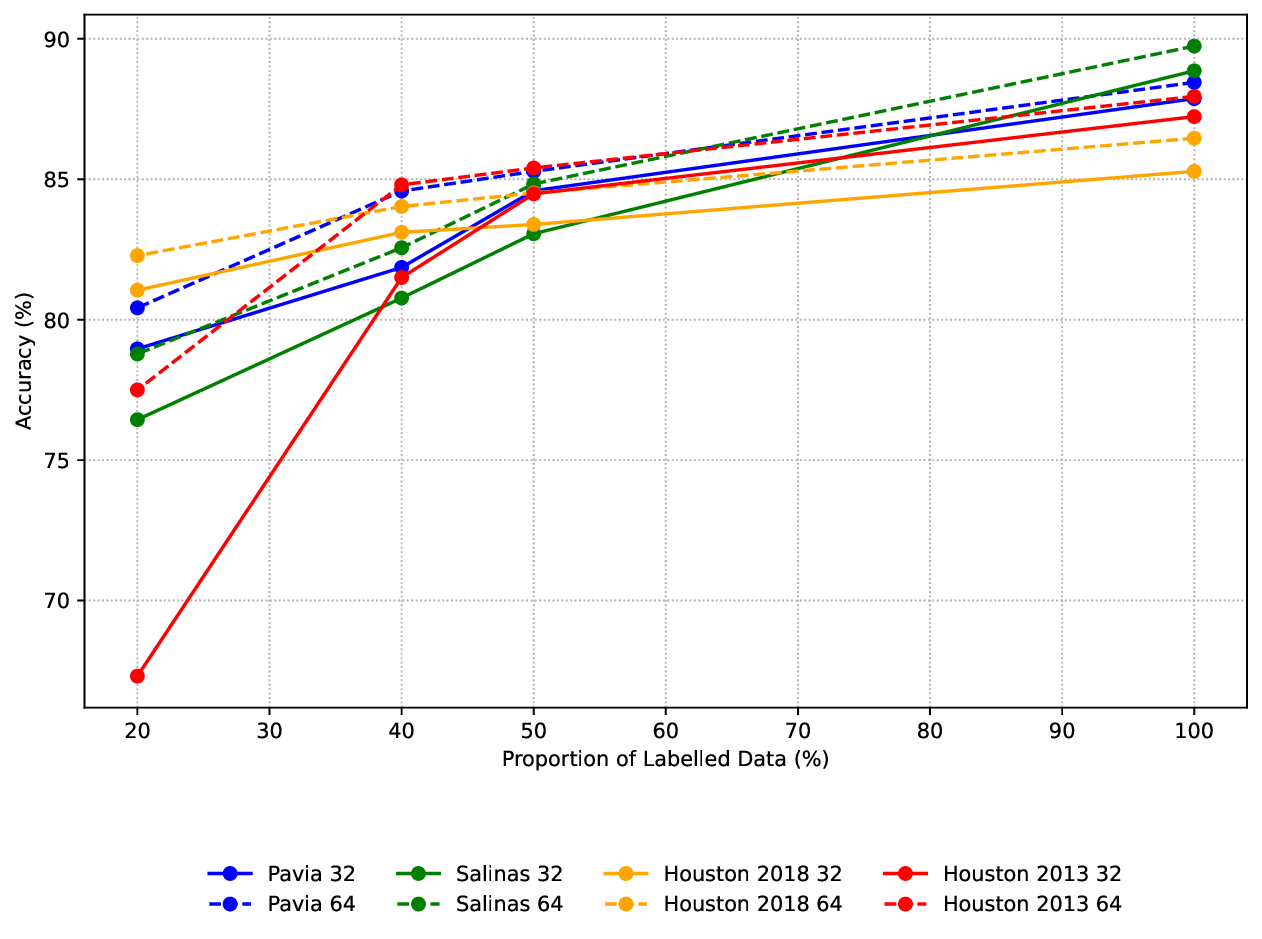}
\caption{CL-based Multi-label Classifier: Model Accuracy Performance (in$\%$)- Impact of Reduced Labelled Data}

\label{fig4:multi-label-impact-data-reduction}
\end{figure}
Figure~\ref{fig4:multi-label-impact-data-reduction} presents the results of testing our \textit{CL-tune} classifier under various data reduction scenarios, $50\%$, $40\%$, and $20\%$. These results encompass two different dimensions of the hidden representation, namely $32$ and $64$. Additionally, we juxtapose these outcomes with the performance of the classifier when trained on the full dataset ($100\%$). 
While reducing the amount of labelled training data predictably leads to a decline in performance, the exact magnitude of this decline varies depending on several factors, which we delineate below.

\noindent\textbf{Optimising the Dimension of Hidden Representations for Enhanced Performance}. We investigated the effect of the hidden representation dimension on the model efficacy across the four datasets under varying levels of data reduction. Notably, with the PaviaU and Houston 2013 datasets, the model's performance exhibited greater resilience to data reduction when a hidden representation of size $64$ was used, compared to size $32$. Figure~\ref{fig4:multi-label-impact-data-reduction} illustrates that models trained on PaviaU and Houston 2013 datasets with a hidden representation of size $64$ (\textit{PaviaU-$64$} and the \textit{Houston 2013-$64$}) were able to withstand up to $60\%$ reduction in training data with only minor decline in performance. In contrast, models using a hidden representation size of $32$ (\textit{PaviaU-$32$} and \textit{Houston 2013-$32$}) experienced a significant performance drop at the same level of data reduction.

For the Salinas dataset, the model with a hidden representation of $64$ (\textit{Salinas-$64$}) consistently outperforms the model with a hidden layer of size $32$ (\textit{Salinas-$32$}), especially as the volume of data for fine-tuning decreases. Compared to the PaviaU and Houston 2013 datasets, changes in hidden representation size have a more pronounced effect on performance for Salinas, with an improvement of ($+0.88\%$) compared to ($+0.58\%$) for PaviaU. However, the performance degradation for Salinas becomes more substantial as the amount of available data is reduced. Given the lower patch variability in the Salinas dataset, even a larger hidden representation size may not suffice to capture subtle feature variations when data is scarce. 
When examining the Houston datasets, we observe similar trends in performance under varying data reduction scenarios and hidden representation sizes. In particular, due to the relatively small size of the original Houston 2013 patches dataset, further reductions beyond $60\%$ result in a significant loss of critical information. This extreme reduction hampers the model's generalisation ability, as the remaining data lacks the diversity and structure necessary for effective training.
Please refer to the Supplementary material document (Section $3$) for a detailed overview of the model performance on all four datasets under varying hidden representation dimensions and data reduction proportions.

These findings accentuate the pivotal role of data characteristics in determining the model's robustness, particularly in scenarios where data availability is limited. While a larger hidden representation size might initially seem beneficial for mitigating the effects of data scarcity, our analyses reveal that the nature of the training data itself is also crucial. The variability and class mixture inherent in the datasets significantly contributes to model performance. For instance, patches from the PaviaU dataset may enhance the model's resilience to data reduction compared to the more uniform and less variable patches in the Salinas dataset. Moreover, a larger hidden representation can help alleviate the negative effects of extreme data reduction by retaining more data characteristics and patterns, compensating for the insufficient training data. This effect is evidenced by the sharp decline in the model's performance on the Houston 2013 dataset when the available data is reduced by more than $60\%$. In this scenario, the smaller hidden representation fails to capture the necessary features for effective learning. 
In conclusion, both hidden representation size and the specific characteristics of the data play instrumental roles in the model's robustness, with neither fully compensating for the limitations of the other. A balanced interplay between these factors is crucial for maintaining model performance, especially when data is scarce. These insights offer valuable guidance for future model development and optimisation in environments where acquiring sufficient training data poses significant challenges.

\noindent\textbf{\textit{CL-tune} versus \textit{CL-freeze}}: 
 An examination of Figure~\ref{fig4:multi-label-impact-data-reduction} and Table~\ref{tab:cl-freeze_vs_cl-tune_acc-h64}, reveals an interesting trend. While the performance of the \textit{CL-freeze} model shows a modest improvement with an increase in hidden representation size for PaviaU and Salinas datasets, it still fails to surpass the performance of the \textit{CL-tune} model. For the Houston 2013 and Houston 2018 datasets, increasing the hidden representation dimension has no positive effect on the \textit{CL-freeze} model; in fact, it leads to a worse performance. This observation aligns with our previous analysis, suggesting that merely increasing the retained information at the hidden representation level does not necessarily enhance the model's performance, particularly when the encoder is not fine-tuned alongside the classifier. Notably, the \textit{CL-tune} model consistently outperforms the \textit{CL-freeze} variant, even when the labelled training data is reduced by $80\%$.
 
\begin{table*}[htbp!]
\begin{center}
\caption{Accuracy performance ($\%$) of the CL-based multi-label classifier with higher-dimensional hidden representation.}
\label{tab:cl-freeze_vs_cl-tune_acc-h64}
 
\setlength\tabcolsep{8pt}  
 
\begin{tabular*}{\textwidth}{@{\extracolsep\fill}lcccc} 
 \toprule
 &{\textbf{\textit{CL-freeze}}}&{\textbf{\textit{CL-tune}}}& {\textbf{\textit{CL-freeze}}}&{\textbf{\textit{CL-tune}}}\\
\midrule
  &\multicolumn{2}{c}{\textbf{\textit{hidden layer $\textbf{32}$}}} & \multicolumn{2}{c}{\textbf{\textit{hidden layer $\textbf{64}$}}}\\
\hline
\textbf{\textit{PaviaU}}&$70.56$&$\textbf{87.87}$&$74.06$&$\textbf{88.45}$\\
\textbf{\textit{Salinas}}&$74.90$&$\textbf{88.86}$&$77.13$&$\textbf{89.74}$\\
\textbf{\textit{Houston 2013}}&$59.84$&$ \textbf{87.23}$&$60.90$&$ \textbf{86.46}$\\
\textbf{\textit{Houston 2018}}&$ 59.27$&$\textbf{85.28}$&$55.96$&$ \textbf{87.95}$\\
\hline
\end{tabular*} 
\end{center}
\end{table*}

The divergence in performance originates from different approaches to handling underlying representations. While the \textit{CL-tune} model continuously refines the representations in conjunction with the classifier, enhancing its adaptability to limited data, \textit{CL-freeze} model locks the encoder's weights, preventing any adaptation of the hidden representations in response to the patterns learned by the classifier. This rigid approach may constrain the model's ability to capture underlying complexities, thereby limiting its performance. In contrast, within a contrastive learning-based framework, the coordinated tuning of the classifier and encoder confers substantial advantages, enabling the model to adapt more effectively to diverse and limited data scenarios.

\noindent\textbf{Impact of Diversity in Data.} 

The PaviaU and Houston 2013 patch datasets, with their diversity and complexity, often exhibit mixed classes. This rich training environment exposes the classifier to a wide range of intrinsic features, thereby alleviating the negative effects of limited labelled data. In stark contrast, the Salinas patch dataset, which is less diverse and more homogeneous, is more vulnerable to the challenges posed by data scarcity. This comparison underscores the significant role that data diversity plays in enhancing classifier performance.

\noindent\textbf{\textit{CL-tune} vs Fully-Supervised Methods: Performance with Reduced Labelled Data.}

Table~\ref{tab:model_acc_less_data_vs_suppervised_learning} presents a quantitative comparison of the performance of \textit{CL-tune} classifier when using only $50\%$ of the training dataset, in comparison to the fully-supervised variants (\textit{Iterative, Joint, and Cascade}) trained on the entire dataset.
With just half of the training data, the \textit{CL-tune} classifier delivers results nearly on par with these fully-supervised counterparts. This finding highlights the strength of the contrastive learning-based encoder in capturing rich data representations, effectively compensating for the limitations posed by a reduced training set.

\begin{table*}[htbp]
\caption{Accuracy performance ($\%$) of the CL-based multi-label classifier with reduced labelled data compared to supervised learning methods trained with full labelled data.}
\label{tab:model_acc_less_data_vs_suppervised_learning}

\setlength\tabcolsep{8pt}  
 
\begin{tabular*}{\textwidth}{@{\extracolsep\fill}lcccc} 
\toprule
& \textbf{\textit{Iterative}} & \textbf{\textit{Joint}} & \textbf{\textit{Cascade}} & \textbf{\textit{CL-tune 50\%}}\\
\midrule
\textbf{\textit{PaviaU}} & $\mathbf84.03$ & $\mathbf86.14$ & $\mathbf83.50$ & $\mathbf84.59$\\
\textbf{\textit{Salinas}} & $\mathbf87.61$ & $\mathbf86.40$ & $\mathbf86.47$ & $\mathbf83.06$\\
\textbf{\textit{Houston 2013}} & $\mathbf 85.81$ & $\mathbf 85.92$ & $\mathbf 56.12$ & $\mathbf84.48$\\
\textbf{\textit{Houston 2018}} & $\mathbf 85.59 $ & $\mathbf 86.03$ & $\mathbf 60.21 $ & $\mathbf 83.39$\\
\hline
\end{tabular*} 
\end{table*}

\subsection{Single-Label Classification}
\label{subsec:single-label-classification}
In this experiment, we investigate several critical aspects of the performance of our classifier, which is trained with a contrastive learning-based encoder, in comparison to classifiers trained using supervised techniques for single-label, patch-level hyperspectral image (HSI) classification tasks. The contrastive learning-based encoder in this experiment employs a hidden representation with a dimensional size of $32$. We compare the performance against methods outlined in the literature, with a primary focus on:

\noindent\textbf{\textit{HSI-CNN}} model as described in~\cite{8455251}. The method extracts spectral-spatial features from the target pixel and its neighbouring pixels, reshapes them into a 2D matrix, and processes this matrix through a standard CNN architecture. Specifically, they use reduced spatial size cubes ($3\times3$) sampled from the original remote sensing datasets, allowing for overlapping regions that maintain a larger number of labelled patches.

\noindent\textbf{\textit{TRI-CNN}} model in~\cite{rs15020316}: This method utilises a three-branch model for hyperspectral image classification. It leverages a multi-scale 3D-CNN architecture to extract spectral and spatial features at multiple scales, which are then combined through feature fusion. Typically, hyperspectral data undergo pre-processing with Principal Component Analysis (PCA) to reduce dimensionality. However, in our experiments, we bypass PCA and test the method using the full extent of the data to maintain consistency with the training conditions of our approach. As a result, we modified the architecture to accommodate the shape and dimensionality of our patch datasets, which we sampled using a stride equal to the patch size to generate non-overlapping patches. 

\noindent\textit{\textbf{SSCL}}: The method described in~\cite{9664575} is a hyperspectral classification approach based on contrastive learning (CL), designed to address the challenge of limited labelled data by leveraging a large number of unlabelled samples during training. The method includes a pre-training phase using contrastive learning, where the model learns to differentiate between positive and negative sample pairs. After this pre-training, the model is fine-tuned on a smaller labelled dataset.
This method utilises the ResNet architecture ~\cite{he2016deep} of varying sizes for the contrastive learning component, mandating larger spatial dimensions such as $21{\times}21$ and $23{\times}23$, in contrast to our $3{\times}3$ spatial format data. To further optimise performance, several data processing techniques are employed:
 
\noindent\textbf{Principal Component Analysis (PCA)} is applied for dimensionality reduction, reducing the depth of the hyperspectral scene to four dimensions to lower the computational complexity.
 
\noindent\textbf{Extended Morphological Profiles (EMP)} are utilised to capture and exploit the spatial correlations within the hyperspectral data.
 
Furthermore, the patch sampling technique used in this method introduces overlapping, leading to significantly larger training and fine-tuning datasets than those used in our approach.

\noindent\textbf{\textit{3D-CNN}} method outlined in~\cite{rs9010067}: This method also leverages spatial context by generating patches of size $5\times5$ with a stride of $1$, resulting in overlapping patches.  The goal is to preserve the original number of labelled samples in the dataset. 
To ensure a fair comparison, we meticulously adapted our architecture and training pipelines to employ the same patch size while reducing the amount of data. Specifically, we followed our approach of generating patches with a stride equal to the patch size—in this case,$5$— which prevents patches from overlapping.
 
Lastly, we also evaluate our method against the three distinct supervised learning training techniques documented in~\cite{rs15245656} and referred to in Section~\ref{subsec:classifier-performance}.
 
This comparative framework not only benchmarks our method against established approaches, allowing for direct comparison, but also highlights its robustness when applied to patch-based datasets. Specifically, it demonstrated the effectiveness of combining spatial and spectral information relative to other architectures that perform similar analysis.

Table~\ref{tab:model_acc_h32_sl} demonstrates the superior performance of our method compared to supervised techniques using $(3{\times}3{\times}b)$ data samples.
In our experiments, the \textit{HSI-CNN} method~\cite{8455251} exhibited reduced performance, likely due to our non-overlapping sampling strategy. This strategy results in fewer training patches compared to the original study, which may have affected the method's effectiveness. Our approach attains higher overall accuracy, highlighting its resilience and efficacy in handling sparsely labelled, high-dimensional data like hyperspectral images. An additional advantage of our method is the absence of pre-processing steps, which significantly reduces both the time and computational complexity. Moreover, avoiding pre-processing minimises the risk of losing valuable information in the original raw data.

\begin{table*}[htbp]  
\caption{Accuracy performance ($\%$) of the CL-based single-label classifier compared to various fully supervised and self-supervised learning schemes.}
\label{tab:model_acc_h32_sl}
 \setlength\tabcolsep{8pt}  
\begin{tabular*}{\textwidth}{@{\extracolsep\fill}lcccccccc} 
\toprule
\textbf{} & \textbf{\textit{Iterative}} & \textbf{\textit{Joint}} & \textbf{\textit{Cascade}} & \textbf{\textit{HSI-CNN}} & \textbf{\textit{TRI-CNN}} & \textbf{\textit{SSCL}} & \textbf{\textit{CL-freeze}} & \textbf{\textit{CL-tune}} \\ 
\midrule
\textbf{\textit{PaviaU}} & $90.71$ & $94.65$ & $87.73$ & $70.30$ & $89.50$ & $88.05$ & $84.02$ & $\textbf{94.87}$ \\ 
\textbf{\textit{Salinas}} & $91.19$ & $93.35$ & $90.34$ & $88.29$ & $90.51$ & $86.78$ & $81.97$ & $\textbf{94.16}$ \\ 
\textbf{\textit{Houston 2013}} & $89.82$ & $\textbf{92.81}$ & $87.82$ & $70.81$ & $80.75$ & $82.37$ & $65.65$ & $91.89$ \\ 
\textbf{\textit{Houston 2018}} & $90.32$ & $93.30$ & $84.68$ & $79.22$ & $90.22$ & $90.41$ & $68.98$ & $\textbf{93.47}$ \\ 
\bottomrule
\end{tabular*}
\end{table*}

\subsubsection{Impact of the Amount of Labelled Data}
\label{subsec: sl-amount-of-data}
In the context of the single-label classification task, Table~\ref{tab:model_acc_Single_label_less_data_vs_supervised_learning} illustrates the impact of a $50\%$ reduction in the labelled data available for training and fine-tuning the contrastive learning-based classifier. While this reduction understandably results in some performance degradation, it is noteworthy that the classifier's effectiveness remains comparable to that of fully supervised methods and self-supervised contrastive learning-based approaches that utilise the full extent of the available datasets. 
 
\begin{table*}[htbp]
\caption{Accuracy performance ($\%$) of the CL-based single-label classifier with reduced labelled data compared to supervised and self-supervised learning methods trained with full labelled data.}
\label{tab:model_acc_Single_label_less_data_vs_supervised_learning}
\setlength\tabcolsep{8pt}  
\begin{tabular*}{\textwidth}{@{\extracolsep\fill}lccccccc} 
 \toprule
&\textbf{\textit{Iterative}}&\textbf{\textit{Joint}}&\textbf{\textit{Cascade}}&\textbf{\textit{HSI-CNN}}&\textbf{\textit{Tri-CNN}}&{\textbf{\textit{SSCL}}}&\textbf{\textit{CL-tune}}\\
\textbf{\textit{Data}}&$100\%$&$\textbf{100}\%$&$100\%$&$100\%$&$100\%$&$100\%$&$\textbf{50}\%$\\
\midrule
\textbf{\textit{PaviaU}}&$90.71$&$\textbf{94.65}$&$87.73$&$70.30$&$89.50$& 88.05&$90.52$\\
\textbf{\textit{Salinas}}&$91.19$&$\textbf{93.35}$&$90.34$&$88.29$&$90.51$&$86.78$&$91.69$\\
\textbf{\textit {Houston 2013}}&$89.82$&$\textbf{92.81}$&$87.82$&$70.81$&$80.75$&82.37&$89.98$\\
\textbf{\textit{Houston 2018}}&$90.32$&$92.30$&$92.30$&$84.68$&$90.22$&90.41&$91.38$\\
\hline
\end{tabular*} 
\end{table*}

\begin{figure}[ht!]
\centering
\includegraphics[width=0.8\linewidth]{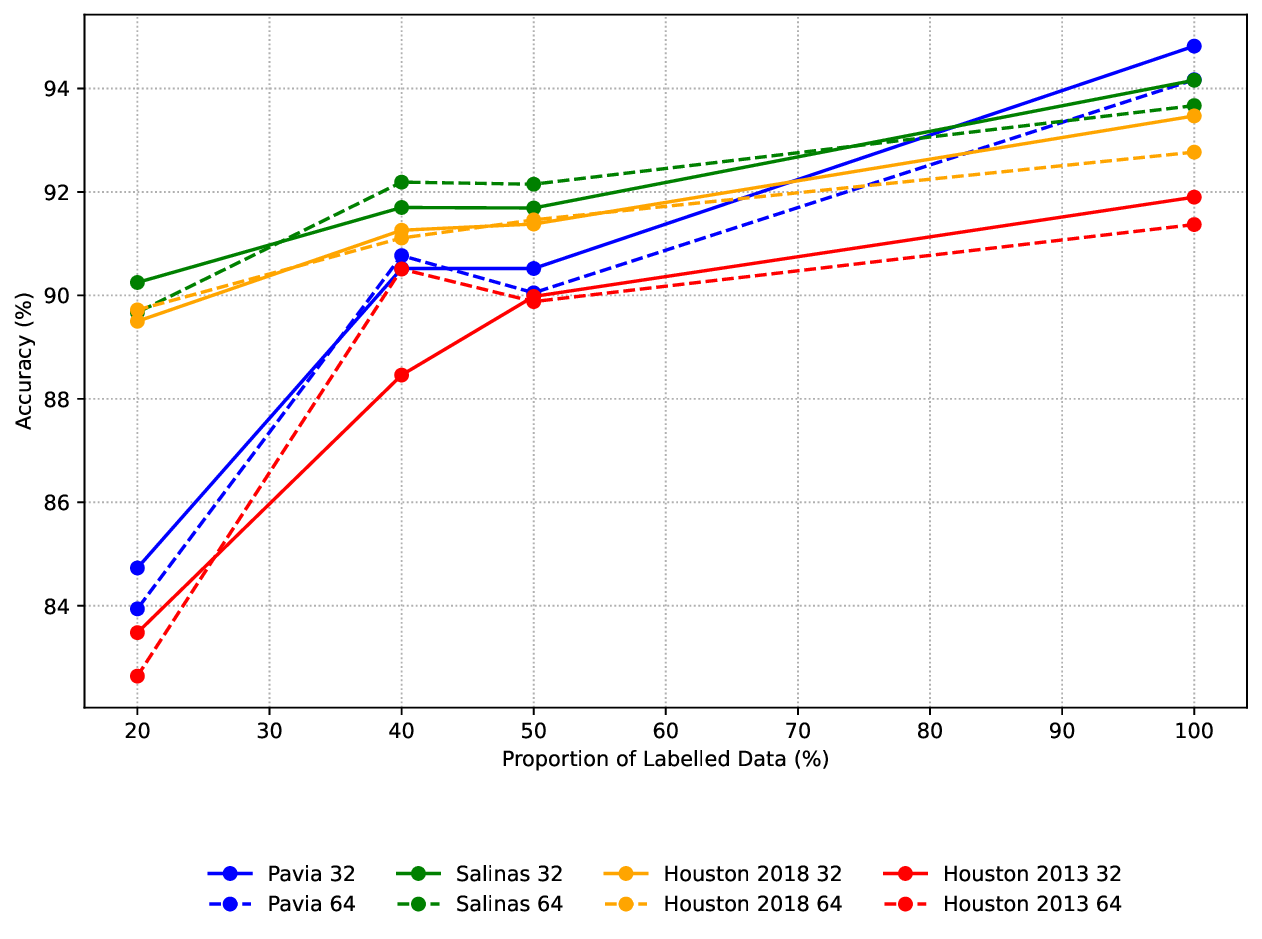}
\caption{CL-based Single-label Classifier: Model Accuracy Performance (in$\%$)- Impact of Reduced Labelled Data}
\label{fig5:single-label-impact-data-reduction}
\end{figure}

Figure~\ref{fig5:single-label-impact-data-reduction} compares the performance of the contrastive learning-based encoder-classifier network across the datasets under four different data reduction proportions and varying hidden representation dimensions. Unlike the multi-label classification scenario, the impact of data reduction on the performance of the contrastive-based encoder-classifier network for single-label data is less pronounced.
Considering the Houston 2013 and PaviaU datasets, we observe that as the available data for fine-tuning the classifier are reduced beyond $50\%$, the classifier struggles to maintain a high level of accuracy. However, with a larger hidden representation, performance remains resilient up to a $60\%$ reduction, followed by a sharp decline. The complexity of the Houston 2013 dataset, characterised by a high rate of mixed patches not adequately described by the available single labels assigned, limits the classifier's ability to extract robust features under data scarcity. 

In the case of PaviaU, the single-label sampling technique excludes patches where the centre pixel is labelled as an unclassified background class. This exclusion reduces the number of mixed patches and may result in a loss of information regarding the correlation between the background and other classes. The absence of this correlation could hinder the classifier's learning process and ultimately limit its overall performance.

For both the Salinas and Houston 2018 datasets, data reduction affects the classifier's performance across different hidden representation sizes, yet the performance remains robust even at the lowest data proportions. This stability arises from the high percentage of uniform patches in these datasets, where each patch is assigned a single label that accurately represents its true class. This uniformity simplifies the classification task, enhancing overall performance.
Please refer to the supplementary material document (Section $3$) for a detailed overview of the model's performance across the four datasets, including the effects of hidden representation dimensions and data reduction proportions.

We also explored the impact of data reduction on the performance of the CL-based classifier by increasing the spatial dimension of the patches to $(5 {\times} 5 {\times} {b})$. This adjustment reduced the total number of data samples available for training. Additionally, we employed a stride of $5$, ensuring no overlap between patches, which further reduced the dataset size. We then used this modified dataset to train the \textit{3D-CNN} method described in~\cite{rs9010067}, which necessitates an identical input shape.

In contrast, the original \textit{3D-CNN} was trained on a more extensive dataset using a patch sampling method that allows overlapping by employing a sliding window that moves one pixel at a time. This technique enables the model to be trained on a significantly larger dataset. Specifically, the \textit{3D-CNN} was trained on $50\%$ of the labelled samples of the PaviaU dataset. Our method, however, adopts a simpler architecture and relies on a smaller dataset - approximately $9\%$ of the total labelled samples for the same dataset- while still achieving superior performance. As shown in Table~\ref{tab:model_acc_h32_sl_5x5_data}, the CL-based method outperforms the \textit{3D-CNN}, especially on PaviaU and Houston 2018 datasets, while nearly matching its performance on the remaining datasets. 
It is worth noting that the \textit{3D-CNN} method was not originally evaluated on the Salinas dataset. In our implementation on this dataset, we adhered to the same assumptions regarding patch size and hyperparameters as outlined in~\cite{deephyperX} for this method.
While the reduction in dataset size may limit the variety of patterns available for the model to learn, our method still demonstrates superior performance, showcasing the effectiveness of our optimised approach in hyperspectral image classification.

\begin{table*}[ht!]
\caption{Accuracy performance ($\%$) of the CL-based single-label classifier compared to supervised learning methods on $(5{\times}5{\times}\text{bands})$ patches.}
 
\label{tab:model_acc_h32_sl_5x5_data}
 
\begin{tabular*}{\textwidth}{@{\extracolsep{\fill}}lll}
\toprule
& \textbf{\textit{3D-CNN}} & \textbf{\textit{CL-tune}} \\ 
\midrule
\textbf{\textit{PaviaU}} & $92.11$ & $\textbf{94.17}$ \\
\textbf{\textit{Salinas}}& $\textbf{94.37}$ & $93.55$ \\
\textbf{\textit{Houston 2013}}& $78.96$ & $78.10$ \\
\textbf{\textit{Houston 2018}}& $88.69$ & $92.63$ \\
\hline
\end{tabular*}  
\end{table*}

\subsubsection{Effect of the Temperature parameter from the Contrastive Learning stage}
\begin{figure}[htpb!]
\centering
\includegraphics[width=0.8\linewidth]{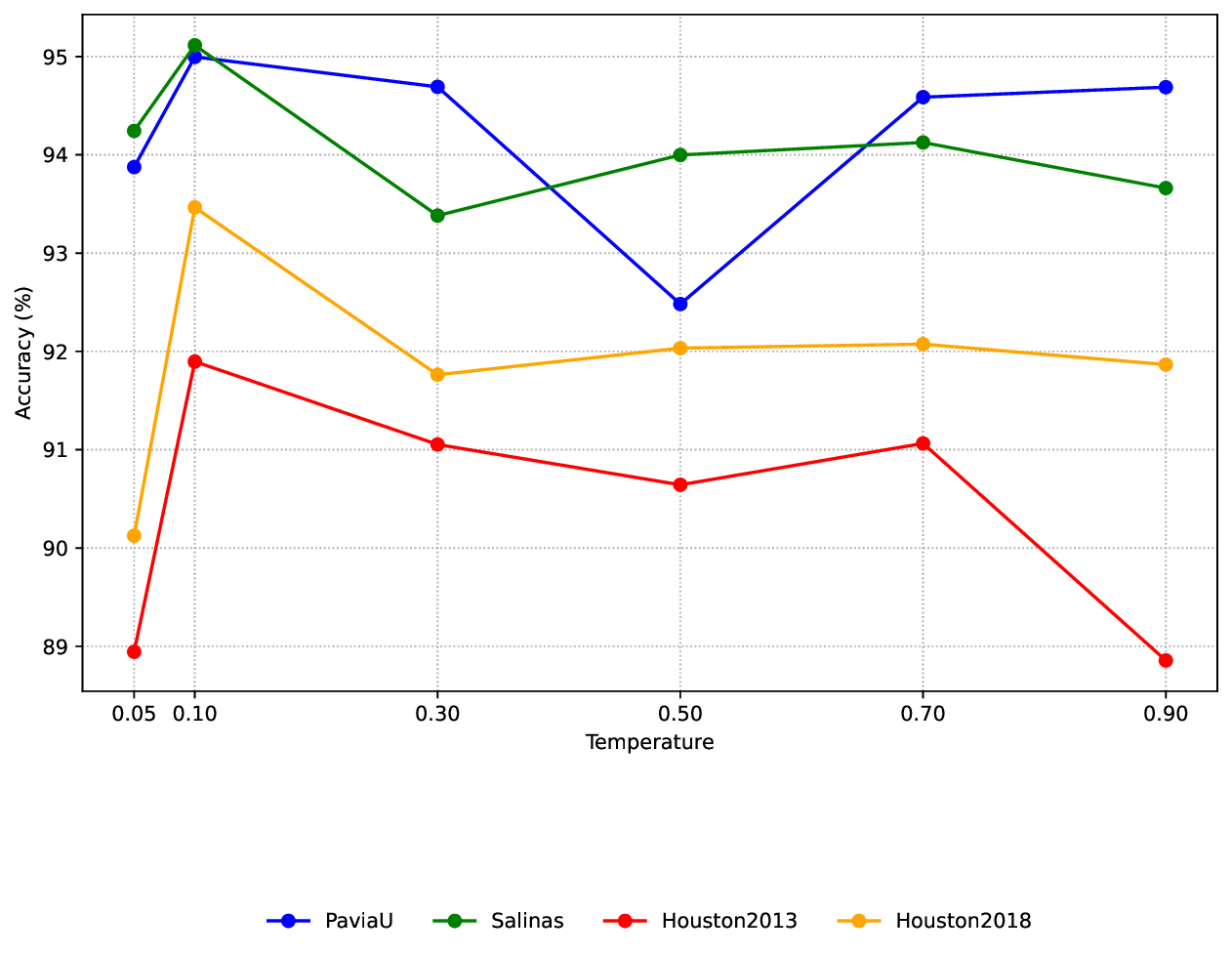}
\caption{Contrastive learning based Single-label Classifier: Model Accuracy Performance (in$\%$)- Impact of different temperature rates}
\label{fig6:single_label_temperature_impact}
\end{figure}

In this experiment, we investigate the suitability of the temperature scaling parameter $T$ used in the Normalised Temperature-scaled Cross Entropy Loss (NT-Xent) contrastive learning loss function~\ref{eq: CLLOSS}. To this end, we retrained the contrastive learning encoder using five candidate values for $T$. Subsequently, we fine-tuned the classifier using each trained encoder and evaluated its performance. Figure~\ref{fig6:single_label_temperature_impact} presents an overview of the classifier's performance across the different $T$ values used during the contrastive learning encoder training. 

The results confirm that $T$$ = 0.1$, used in our initial training, is the optimal value, yielding the highest accuracy. We conclude that a temperature parameter of $T$$ = 0.1$ achieves an optimal balance between model confidence and generalisation. At this value, the model effectively distinguishes positive pairs with high confidence while still appropriately accounting for negative pairs.

\section{Qualitative Analysis of the Hidden Representations}
\label{sec:t-sne visualisation}

In this section, we conduct a qualitative analysis with two primary objectives. First, we explore the hidden representations learned by the encoder within a contrastive learning framework. Second, we compare these representations to those derived from a two-component network comprising an encoder and a classifier, trained jointly in a supervised manner.
To achieve these objectives, we utilise t-distributed Stochastic Neighbour Embedding (t-SNE) to visualise the high-dimensional data. Each point in the t-SNE visualisation represents a high-dimensional feature vector reduced to a two-dimensional space. The colour and shape of each point correspond to a specific class label, as indicated in the legend.
We apply the t-SNE visualisation technique to the hidden representations extracted from the feature extraction component under the two training methods described above. 
First, we examine the representations generated by the contrastive learning-based encoder trained in conjunction with a classifier. Second, we investigate the representations learned by the encoder as part of the two-component network of autoencoder and classifier trained using the \textit{Joint} training scheme. Our investigation encompasses both the multi-label and the single-label prediction tasks. 
For this analysis, we focus on the PaviaU and Salinas public datasets,  which are widely used in hyperspectral imaging research. These datasets allow us to gain insights into the feature spaces that the models have assimilated. To ensure a fair comparison between the two methods, we set the size of the hidden representations to $32$. 

\subsection{t-SNE Visualisation: Multi-label Patches}
\begin{figure}[htbp]
    \centering
    \begin{subfigure}[t]{0.48\textwidth}  
    \centering 
    \includegraphics[width=\textwidth]{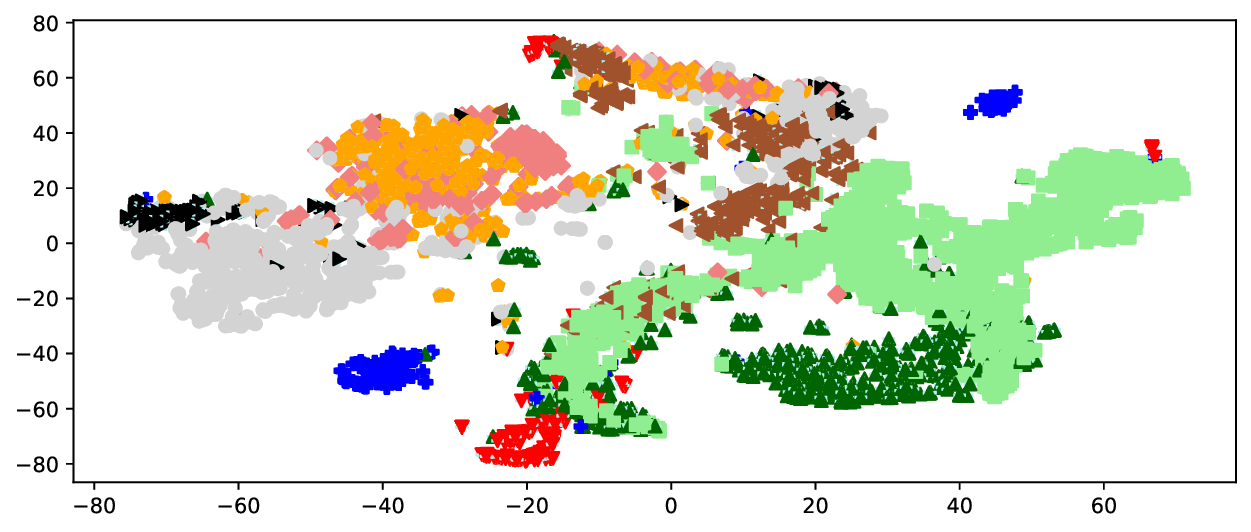} 
    \caption{} 
    \label{fig:sub4_a_paviau_cl_classifier_multi} 
    \end{subfigure}
     \begin{subfigure}[t]{0.48\textwidth} 
    \centering 
  
    \includegraphics[width=\textwidth]{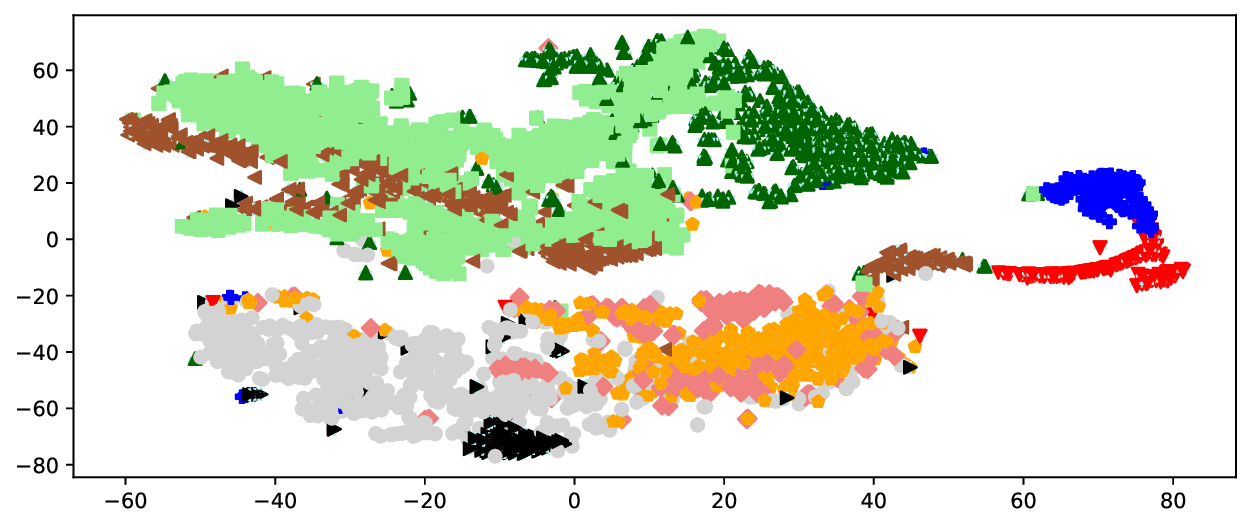}
    \caption{}
    \label{fig:sub4_b_paviau_joint_multi} 
    \end{subfigure}
   
    \begin{subfigure}{0.75\textwidth}
    \centering 
        \includegraphics[height=1.5cm, width=\textwidth]{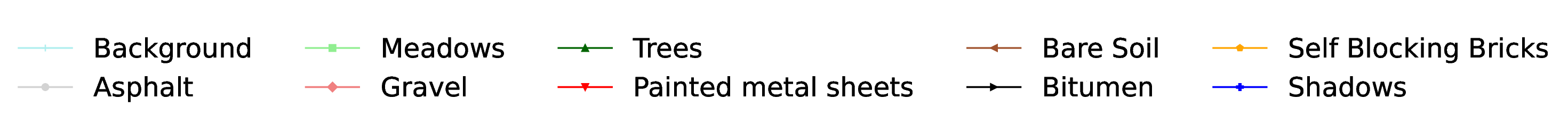}
    \label{fig:sub4_c_tsne_legend_multi} 
    \end{subfigure}
    
    \caption{PaviaU: Contrastive Learning (a) vs. Supervised Joint Training (b) - Multi-label Classification.}
    \label{fig7:tsne_paviau_multi_label}  
\end{figure}

Figure~\ref{fig7:tsne_paviau_multi_label} exhibits the distribution of the classes in a low-dimensional space of the hidden representations under the two scenarios described earlier. In Figure~\ref{fig:sub4_a_paviau_cl_classifier_multi}, the representations are extracted from the classifier network fine-tuned with the contrastive learning-based encoder.The visualisation reveals that classes are well-defined based on the nature of the materials they represent. 
In the context of remote sensing, classes such as \enquote{Trees}, \enquote{Meadows}, and \enquote{Bare Soil} are natural land cover types that often coexist and exhibit similar spectral signatures, as they are vegetative or natural ground materials. These similarities in their high-dimensional input space cause these classes to cluster in proximity when visualised in two dimensions using t-SNE. Similarly, \enquote{Asphalt}, \enquote{Gravel}, and \enquote{Bitumen} are all human-made surfaces commonly used in construction and paving. They reflect light in similar ways, leading to comparable spectral characteristics. The encoder captures these similarities, leading to the clustering of these classes in the t-SNE visualisation.

While this separation among classes  suggests that the model has effectively learned to distinguish certain features, it might still be conflating features within classes of a similar nature, as evidenced by the overlapping areas we observe in Figure~\ref{fig7:tsne_paviau_multi_label}. This overlap can also be attributed  to the multi-labelled nature of the data, where instances contain combinations of classes. 
The t-SNE visualisation reveals an intriguing clustering phenomenon, highlighting the model's nuanced capability to discern patterns within the data. 
Notably, despite the absence of explicit geographic information during the training of the encoder or the fine-tuning of the encoder-classifier network, the representations exhibit a form of spatial intelligence. For example, classes like \enquote{Shadow} and \enquote{Painted metal sheets} are dispersed across different regions in the visualised space, reflecting transitional properties or attributes change gradually and continuously,  possibly resonating with geographical proximity. These observations suggest that the model has inherently captured contextual and transitional features of the classes, likely as a byproduct of the contrastive learning method.  For instance, \enquote{Shadows} may appear differently depending on the objects casting them, their environment, or the time of day, while \enquote{Painted metal sheets} could vary in appearance due to factors like wear, rust, or the angle at which they are viewed. 
Contrastive learning contributes to this effect, as it significantly enhances the encoder's ability to capture invariant representations by focusing on the inherent properties of the data. 
This approach facilitates the recognition of similarities and differences between examples, emphasising the the model's ability to identify stable, intrinsic characteristics across multiple views of the data. By strategically employing data augmentation to create diverse perspectives, contrastive learning strengthens the model's capacity to discern these core features. Ultimately, this methodology fosters a deeper understanding of the data, leading to more robust and invariant feature recognition.
In our analysis, contrastive learning enables the model to learn representations that naturally encode notions of transitional properties or adjacency, even though these aspects are not explicitly part of the training signal. This emergent clustering, which aligns with natural or human-made categories, is especially valuable in remote sensing applications, where recognising context-dependent features can enhance predictive performance.
Figure~\ref{fig:sub4_b_paviau_joint_multi} demonstrates the capacity of the jointly trained network to separate high-dimensional data into distinct clusters, suggesting effective feature representation. This distinct clustering becomes particularly insightful when considering that the jointly trained model optimises a loss function significantly weighted towards the input reconstruction rather than classification. This prioritisation is evident in the clustering of features within the plot, indicating that the autoencoder has successfully learned generalised features that help to reconstruct the input data with fidelity. While the weighted loss function underlying the training process of this technique suggests a lower emphasis on classification, the differentiation of classes into separate clusters indicates that the model has incidentally learned to distinguish between classes to a certain extent. 
However, the bias towards reconstruction has contributed to the overlapping between some classes, notably \enquote{Bare Soil} and \enquote{Meadows}. This overlap underscores that the classification aspect is not the training priority and leaves room for improvement. In contrast to the classifier network fine-tuned with the contrastive learning-based encoder, which tends to capture transitional and context-related features, this network has failed to detect these aspects. This oversight results from the network's tendency to prioritise capturing features that will assist the autoencoder in reconstruction over discriminative features that will improve the performance of the multi-label classifier. Consequently, while the network excels in reconstruction, its ability to classify is relatively underdeveloped and could benefit from a rebalanced training objective that gives more weight to classification accuracy. 
\begin{figure}[htbp]
    \centering
    \begin{subfigure}[t]{0.48\textwidth} 
        \centering 
        \includegraphics[width=\textwidth]{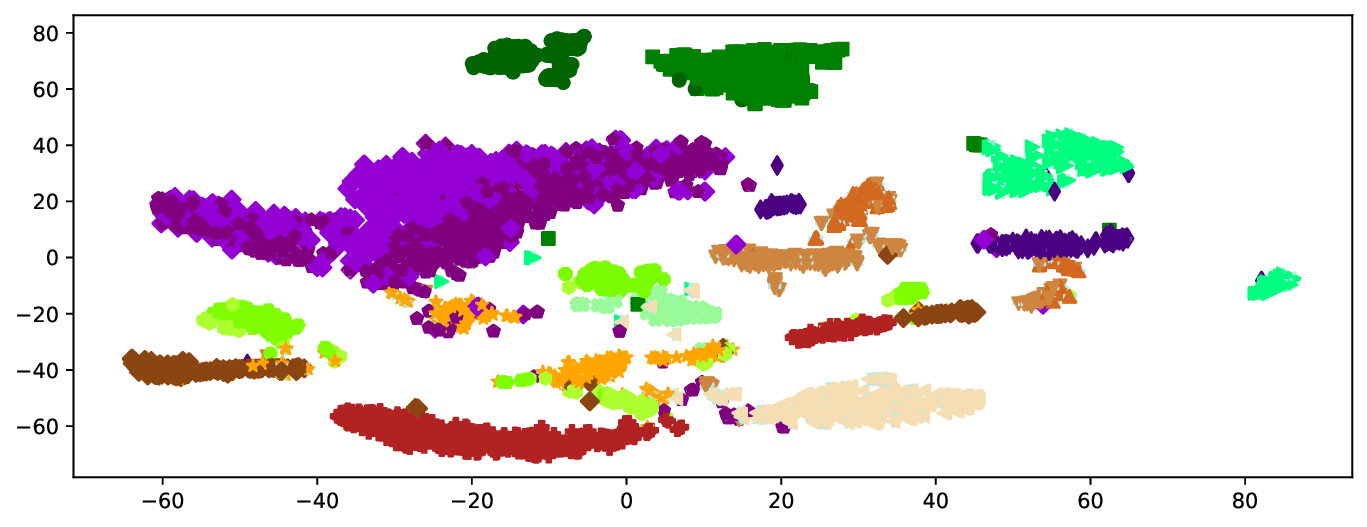}
        \caption{} 
        \label{fig:8_a_salinas_cl_classifier_multi}
    \end{subfigure}
    \hfill
    \begin{subfigure}[t]{0.51\textwidth}
        \centering  
        \includegraphics[width=\textwidth]{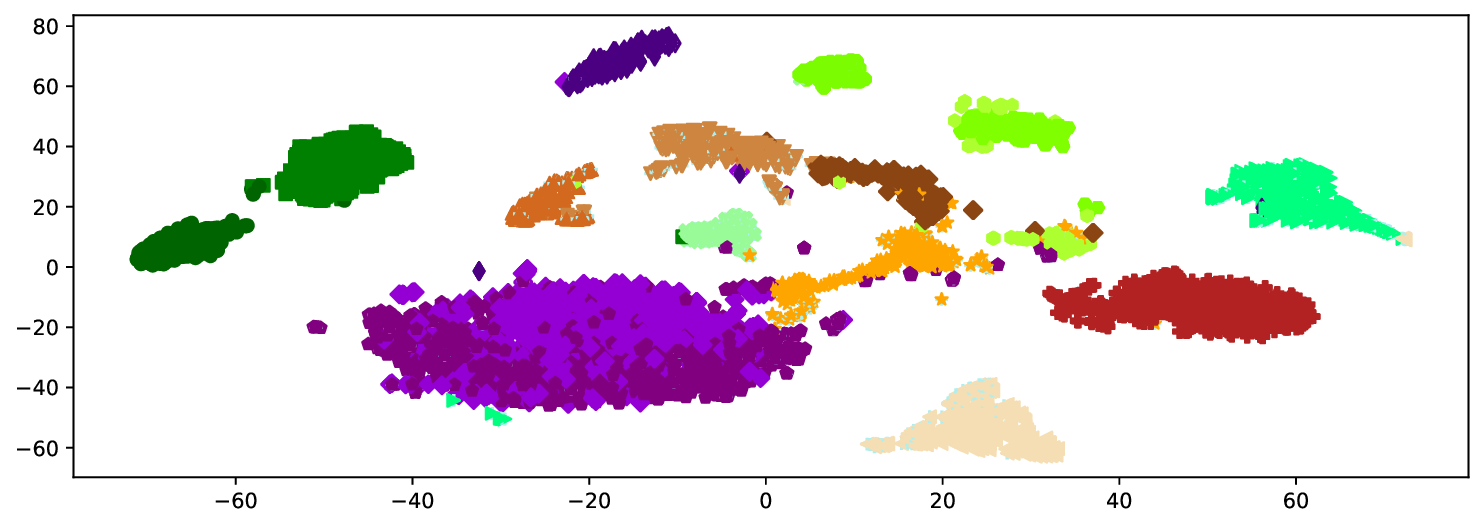}
        \caption{} 
        \label{fig:8_b_salinas_joint_multi}
    \end{subfigure}
    \begin{subfigure}[b]{0.75\textwidth}
        \centering 
        \includegraphics[height=2.5cm, width=\textwidth] {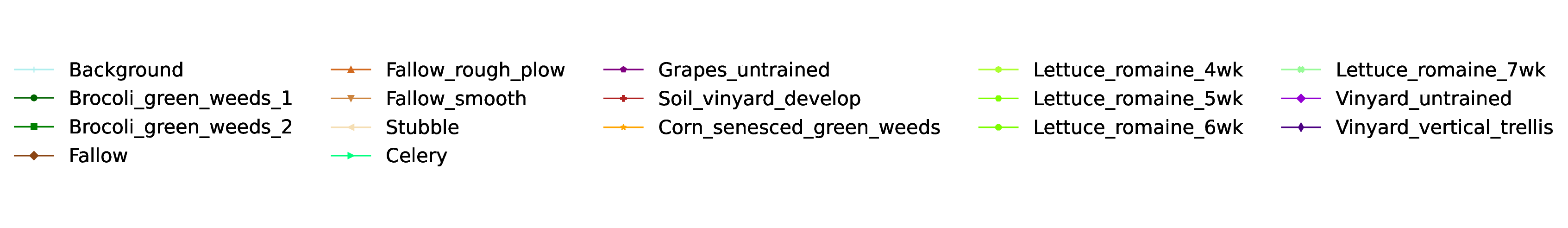}
        \caption*{}  
        \label{fig:8_c_tsne_legend_multi} 
    \end{subfigure}

    \caption{Salinas: Contrastive Learning (a) vs. Supervised Joint Training (b) - Multi-label Classification.}
    \label{fig8:8_tsne_salina_multi_label}
\end{figure}

In Figure~\ref{fig8:8_tsne_salina_multi_label}, we observe similar grouping patterns using the Salinas dataset despite the inherent difference in the nature of patches between the Salinas and PaviaU datasets. The Salinas dataset predominantly consists of uniform patches, unlike PaviaU. The joint training scheme (Figure~\ref{fig:8_b_salinas_joint_multi}) of the two-component network has adeptly captured the essential features for reconstruction, leading to a near-perfect class segregation in the low dimensional t-sne visualisation. This clear clustering facilitates class differentiation by the classifier. However, the uniformity in the patches extracted from the Salinas dataset limits the emergence of any transitional and context-driven features, which are crucial for the robustness and generalisation of the classifier. Yet, the classifier fine-tuned with a contrastive learning-based encoder has successfully identified those intricate relationships despite the scarcity of mixed patches. The proximity and occasional overlap between classes in Figure~\ref{fig:8_a_salinas_cl_classifier_multi} exemplify this phenomenon.

\subsection{t-SNE Visualisation:Single-label Patches}

\begin{figure}[htbp]
    \centering
    \begin{subfigure}[t]{0.49\textwidth}
    \centering
    \includegraphics[width=\textwidth]{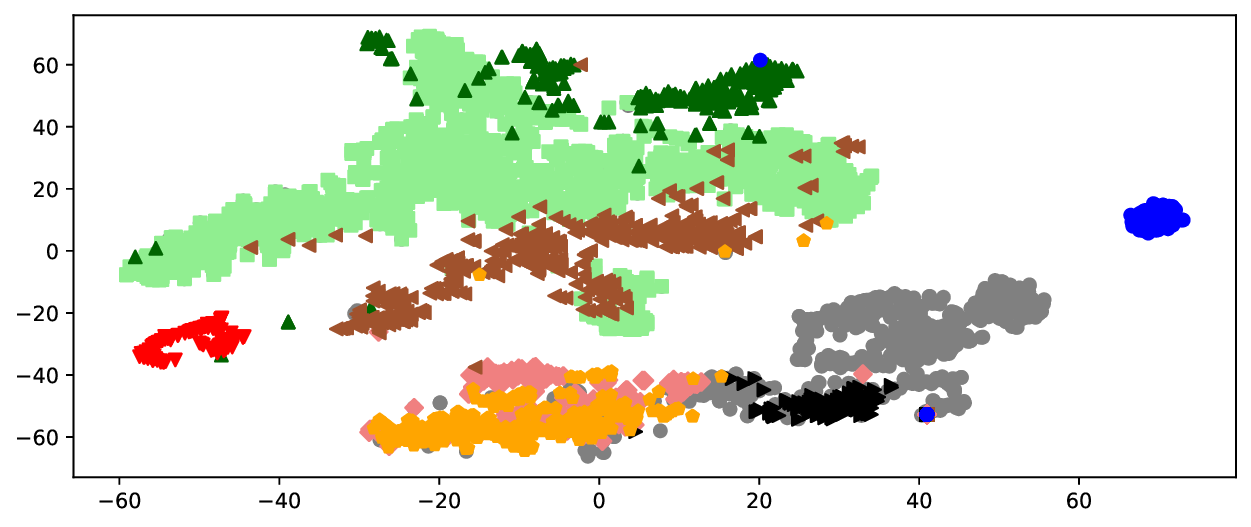}
    \caption{} 
    \label{fig:9_a_paviau_contrast_classifier_single}
    \end{subfigure}
    \hfill
    \begin{subfigure}[t]{0.49\textwidth}
    \centering
    \includegraphics[width=\textwidth]{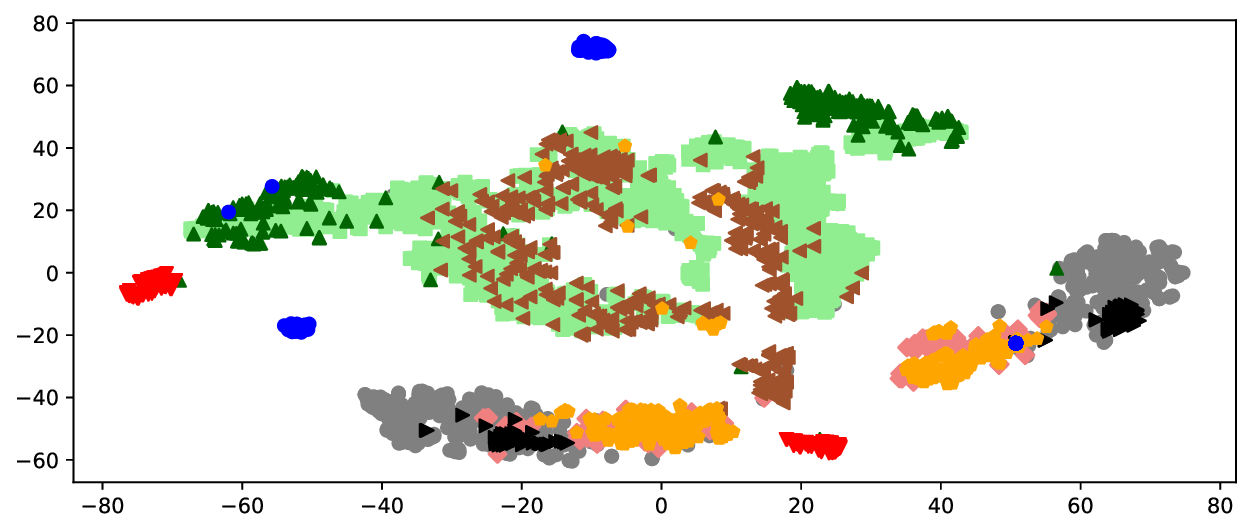}
    \caption{}  
    \label{fig:9_b_paviau_joint_single}
    \end{subfigure}
    \begin{subfigure}{0.75\textwidth}
    \centering
    \includegraphics[height=1.5cm, width=\textwidth]{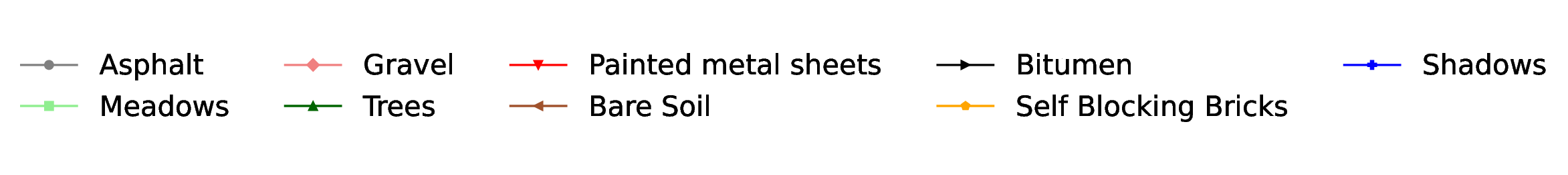}
    \caption*{}  
    \label{fig:9_c_tsne_legend_single}  
    \end{subfigure}
    \caption{PaviaU: Contrastive Learning (a) vs. Supervised Joint Training (b)- Single label Classification.}
    \label{fig9:9_tsne_paviau_single}
\end{figure}

Figure~\ref{fig9:9_tsne_paviau_single} illustrates the hidden representation space of PaviaU patches with single labels, generated by both the contrastive learning-based encoder fine-tuned with a classifier and the encoder within a jointly trained autoencoder-classifier network. In the contrastive learning scenario (Figure~\ref{fig:9_a_paviau_contrast_classifier_single}), the encoder effectively captures features related to the compositional characteristics of the classes, as evidenced by the distinct groupings in the low-dimension feature space. However, it is less influenced by the contextual and transitional characteristics of the classes. The contrastive learning technique leads to a feature space where \enquote{Meadows}, \enquote{Trees}, and \enquote{Soil} are closely positioned, indicating similar feature representations related to the natural characteristics of their material. In contrast, human-made materials such as \enquote{Asphalt}, \enquote{Gravel}, and \enquote{Bitumen} form separate groupings. \enquote{Shadow} and \enquote{Painted metal sheets} each occupy distinct positions in the space, highlighting their unique features. 

These patterns suggest that the hidden representations generated by the contrastive learning based encoder under a single-label context, provide the classifier with robust discriminative features. This is a particularly promising observation, as it demonstrates that the encoder can produce strong features, even when prior-knowledge is limited to a single label corresponding to the centre pixel of the patch.

However, the differences observed when compared to the visualisation under the multi-label scenario in Figure~\ref{fig:sub4_a_paviau_cl_classifier_multi} suggest that the classifier's lack of access to multi-label contextual information may explain the absence of contextual connections between classes.

In contrast, the feature space resulting from the joint training approach exhibits a more blended spatial relationship among classes, even within the single-label context. While \enquote{Trees}, \enquote{Meadows}, and \enquote{Soil} still cluster closely, and \enquote{Asphalt}, \enquote{Gravel}, and \enquote{Bitumen} form another grouping, there is less distinction between classes compared to the contrastive learning technique.
Interestingly, the \enquote{Soil} class appears isolated in a separate region, and while \enquote{Shadows} and \enquote{Painted metal sheets} remain distinguishable from other classes, they are divided into two subgroups. The joint training approach, which aims to enhance the encoder's ability to produce general features conducive to input reconstruction, yields hidden representations that are less class-specific and more generalised. While this is beneficial for generative tasks, it may weaken the sharp class boundaries critical for discriminative classification. Notably, Figure~\ref{fig:9_b_paviau_joint_single} raises intriguing questions about the spatial distribution of class representations. The occurrence of the same class in disparate regions of the feature space, such as \enquote{Shadows} and \enquote{Painted metal sheets}, is an unexpected phenomenon that merits further investigation. It prompts a discussion on the consistency of feature representation and the factors contributing to this variability within the same class.

The contrastive learning technique, in contrast, demonstrates a clearer distinction between class groupings, where compositional similarities are preserved while maintaining greater inter-class separation. 
The differences in class representations between the two approaches highlight a complex interplay between the encoding strategy and the label context, prompting further exploration into the consistency of feature representation and its impact on classification performance in hyperspectral imaging.
\begin{figure}[htbp]
    \centering
    \begin{subfigure}[t]{0.47\textwidth}
    \centering
    
    \includegraphics[width=\textwidth] {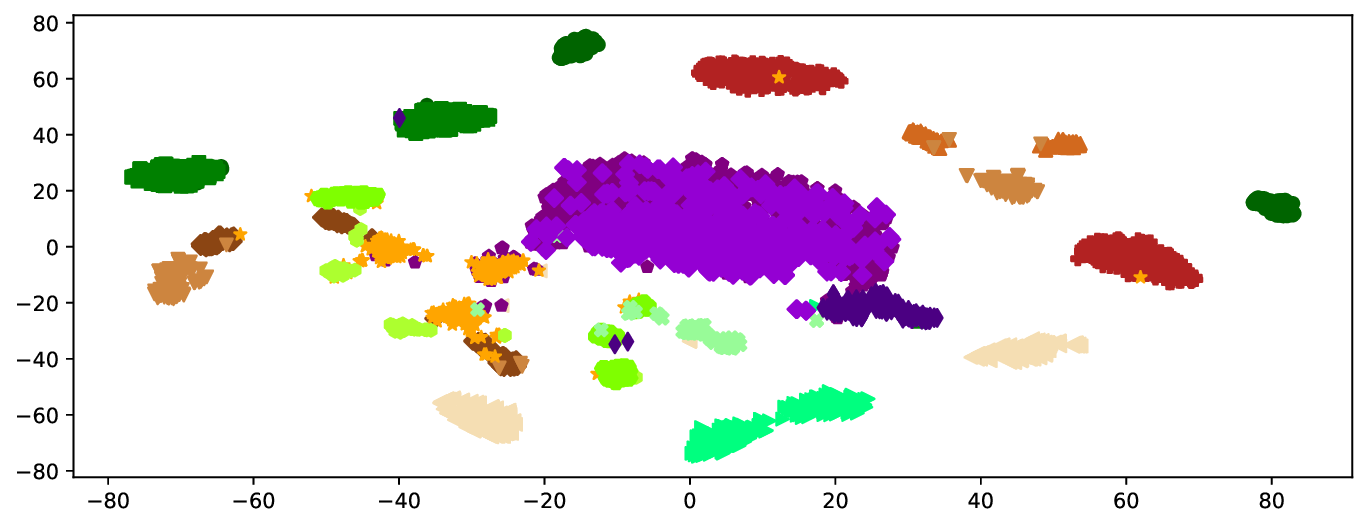}
    \caption{}  
    \label{fig:10_a_salinas_contrast_classifier_single}
    \end{subfigure}
    \hfill
    \begin{subfigure}[t]{0.51\textwidth}
    \centering
    
    \includegraphics[width=\textwidth]{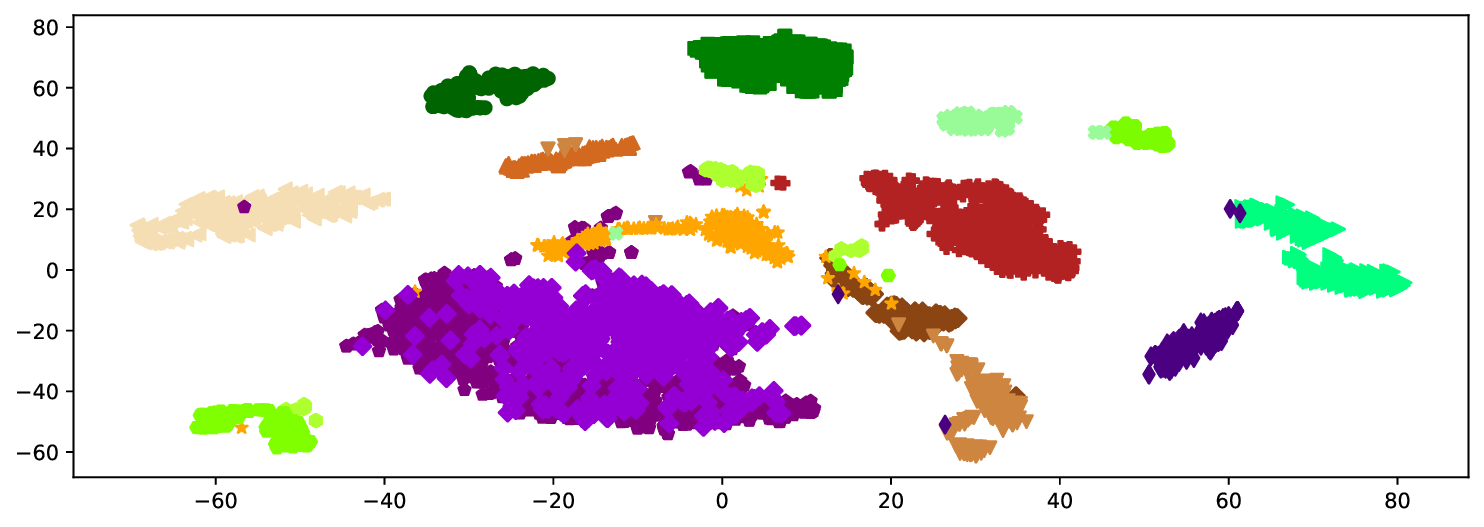}
    \caption{}  
    \label{fig:10_b_salinas_joint_single}
    \end{subfigure}
  
    \begin{subfigure}{0.75\textwidth}
    \centering
    \includegraphics[height=2cm,width=\textwidth]{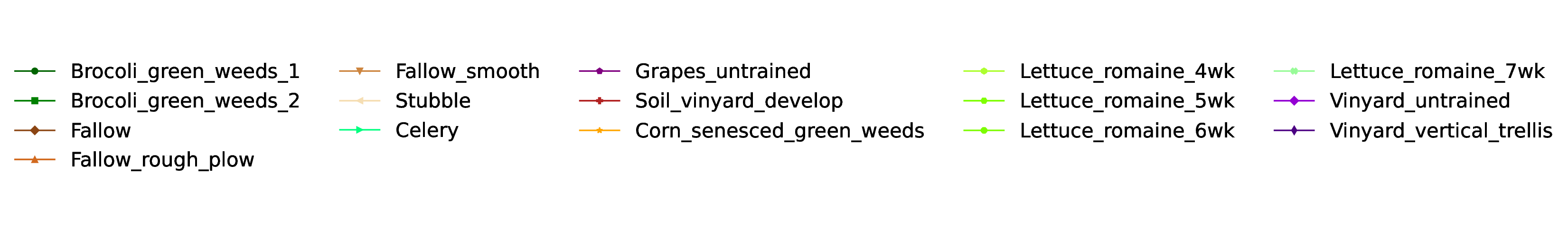}
    \caption*{} 
    \label{fig:10_c_tsne_legend_single} 
    \end{subfigure}

    \caption{Salinas: Contrastive Learning (a) vs. Supervised Joint Training (b) - Single label Classification.}
    \label{fig10:10_tsne_salinas_single_label}
\end{figure}

In Figure~\ref{fig10:10_tsne_salinas_single_label}, we present the hidden representations of the Salinas dataset within the single-label context, generated by both the contrastive learning-based encoder and the encoder component of the jointly trained two-component network. 
Figure~\ref{fig:10_a_salinas_contrast_classifier_single} illustrates that the hidden representations produced by the contrastive learning-based encoder tend to capture contextual features. A key challenge in the Salinas dataset is that many classes belong to vegetation and earth-like categories characterised by common nuances and potentially overlapping spectral signatures. Nevertheless, we observe meaningful groupings, such as \enquote{Soil} and \enquote{Fallow}; \enquote{Grapes}, and both classes of \enquote{Vineyards}; \enquote{Broccoli$\_$1} and \enquote{Broccoli$\_2$}; \enquote{Celery} and various types of \enquote{Lettuce}. Interestingly, the encoder also seems to capture certain spatial aspects from the original scene, even though this information was not part of the training data. For instance, \enquote{Corn} and \enquote{Lettuce} appear closely positioned in the feature space, as do \enquote{Fallow} and \enquote{Fallow smooth}, and \enquote{Stubble} and \enquote{Celery}, which is consistent with their geographical proximity in the original scene.
Turning to Figure~\ref{fig:10_b_salinas_joint_single}, it shows that the encoder from the jointly trained network succeeds in creating distinct groupings of classes, irrespective of compositional similarity or geographical proximity. As with previous scenarios, the joint encoder emphasises general features to facilitate the reconstruction process. This results in class clusters that enable the classifier to capture distinctive features crucial for its task.
An additional observation is the presence of more outliers in the representations produced by the jointly trained encoder compared to the contrastive learning-based encoder. This observation indicates that while the joint encoder focuses on capturing broader, generalisable features, it may also introduce greater variability in the hidden representations.

\section{Conclusions}
\label{sec:conclusion}
This paper introduces a two-stage contrastive-learning-based classification method specifically designed for hyperspectral remote sensing images. Our approach utilises a minimalist architecture that is significantly enhanced by self-supervised contrastive learning. A key strength of this model lies in its robustness in scenarios characterised by a scarcity of labelled training data. It demonstrates successful application in both multi-label and single-label classification tasks. By reducing reliance on extensive labelled datasets, this contrastive learning-based model offers a viable solution for HSI analysis, particularly in data-limited contexts. We provide a qualitative analysis using low-dimensional t-SNE visualisations to illustrate the enhanced capabilities of our contrastive learning-based encoder.
This analysis demonstrates how the contrastive learning approach enriches data representations by integrating contextual and geographical information into the hidden representations it generates, despite these signals being absent during training. This enrichment underscores the model's ability to generalise effectively and leverage inherent data structures without direct supervision.

Looking ahead, our research will encompass several key directions. First, we will expand our empirical validation by incorporating a broader range of hyperspectral datasets to assess the generalisability and robustness of our method across diverse data sources. Second, we plan to investigate the role of attention-based networks in hyperspectral data analysis, exploring how these architectures can enhance feature representation and model performance. Finally, we aim to focus on model interpretability to deepen our understanding of the decision-making processes within these complex systems and improve transparency—an essential aspect for building trust and enabling practical applications.

\section*{Acknowledgments}
This work is supported by Flanders Innovation $\&$ Entrepreneurship-VLAIO, under grant no. HBC.2020.2266.


\section{Supplementary Material}
\subsection{Introduction}
\label{sec:sup-mat-intro}
The supplementary material provides additional details that further elucidate various aspects of the method's training process, alongside comprehensive performance results of data reduction scenarios. Specifically, it offers a detailed account of the method's accuracy under different scenarios involving varying percentages of reduced labeled training data. Section~\ref{sec: hyperparameters} presents the various hyperparameters applied for training on the four different datasets, while section ~\ref{sec:data_reduction_results} presents the results for each dataset, under both hidden representation scenarios. These provide further insight into Sections~\ref{subsec: ml-amount-of-data} and \ref{subsec: sl-amount-of-data}of the main manuscript focusing on the impact of reduced labelled training data.  
 
\subsection{Training Parameters}
\label{sec: hyperparameters}
This section provides a comprehensive overview of all hyperparameters used for training the contrastive learning-based encoders as well as the classifier models, across both multi-label and single-label classification tasks. The hyperparameters include learning rates, batch sizes, number of epochs, and regularization parameters, among others. By offering this level of detail, the intention is to ensure that other researchers have sufficient information to accurately reproduce the methodology and results presented in this work. Reproducibility is a cornerstone of scientific research, and by specifying all relevant training configurations, including both dataset-specific and general settings, this work aims to facilitate the validation and extension of the proposed approach by the broader research community.

\begin{table*}[htbp!]
\centering
\caption{Hyperparameters for the Multi-label Contrastive Learning-based Encoder Model}
\label{tab:multi-label-contrastive_learning_hyperparameters}

\begin{tabular*}{\textwidth}{@{\extracolsep{\fill}}lllll}
\toprule
\textbf{Parameter}                & \textbf{PaviaU} & \textbf{Salinas} & \textbf{Houston 2013} & \textbf{Houston 2018} \\ \midrule
\textbf{Epochs}                   & 85              & 85               & 50                    & 50                    \\
\textbf{Batch Size}               & 300             & 300              & 64                    & 64                    \\
\textbf{Learning Rate}            & $1 \times 10^{-3}$  & $1 \times 10^{-2}$  & $1 \times 10^{-3}$  & $1 \times 10^{-3}$      \\
\textbf{Gamma}                    & 0.9             & 0.9              & 0.9                   & 0.9                   \\
\textbf{Lr-Step Size}             & 10              & 7                & 10                    & 10                    \\
\textbf{L2-Regularization Weight} & $1 \times 10^{-4}$  & $1 \times 10^{-5}$  & $1 \times 10^{-4}$  & $1 \times 10^{-4}$      \\
\textbf{Temperature}              & 0.1             & 0.1              & 0.1                   & 0.1                   \\
\textbf{Dropout}                  & 0.3             & 0.3              & 0.5                   & 0.5                   \\
\bottomrule
\end{tabular*}
\end{table*}

\begin{table*}[htbp!]
\centering
\caption{Hyperparameters for the Single-Label Contrastive Learning-based Encoder Model}
\label{tab:single_label_contrastive_learning_hyperparameters}
 
\begin{tabular*}{\textwidth}{@{\extracolsep{\fill}}lllll}
\textbf{Parameter}                & \textbf{PaviaU} & \textbf{Salinas} & \textbf{Houston 2013} & \textbf{Houston 2018} \\ \midrule
\textbf{Epochs}                   & 50              & 50               & 50                    & 50                    \\
\textbf{Batch Size}               & 400             & 300              & 64                    & 120                   \\
\textbf{Learning Rate}            & $1 \times 10^{-2}$  & $1 \times 10^{-2}$  & $1 \times 10^{-3}$  & $1 \times 10^{-3}$      \\
\textbf{Gamma}                    & 10              & 10               & 10                    & 10                    \\
\textbf{Lr-Step Size}             & 0.9             & 0.9              & 0.9                   & 0.9                   \\
\textbf{L2-Regularization Weight} & 0               & 0                & $1 \times 10^{-4}$  & $1 \times 10^{-4}$      \\
\textbf{Temperature}              & 0.1             & 0.1              & 0.1                   & 0.1                   \\
\textbf{Dropout}                  & 0.6             & 0.3              & 0.5                   & 0.3                   \\
\bottomrule
\end{tabular*} 
\end{table*}

\begin{table*}[htbp!]
\centering
\caption{Hyperparameters for the Multi-Label Contrastive-Learning based Classification Model}
\label{tab:multi-label-classification_hyperparameters}

\begin{tabular*}{\textwidth}{@{\extracolsep{\fill}}lllll}
\toprule
\textbf{Parameter}                & \textbf{PaviaU} & \textbf{Salinas} & \textbf{Houston 2013} & \textbf{Houston 2018} \\ \midrule
\textbf{Epochs}                   & 256             & 200              & 450                   & 200                   \\
\textbf{Batch Size}               & 260             & 164              & 16                    & 16                    \\
\textbf{Learning Rate}            & $1 \times 10^{-3}$  & $1 \times 10^{-3}$  & $1 \times 10^{-3}$  & $1 \times 10^{-3}$      \\
\textbf{Gamma}                    & 0.9             & 0.9              & 0.6                   & 0.9                   \\
\textbf{Lr-Step Size}             & 10              & 10               & 20                    & 20                    \\
\textbf{L2-Regularization Weight} & $1 \times 10^{-4}$  & $1 \times 10^{-4}$  & $5 \times 10^{-4}$  & $9 \times 10^{-6}$      \\
\textbf{Dropout (Classifier)}     & 0.6             & 0.6              & 0.3                   & 0.2                   \\
\textbf{Dropout (Encoder)}        & 0.3             & 0.3              & 0.3                   & 0.2                   \\
\bottomrule
\end{tabular*} 
\end{table*}

\begin{table*}[htbp!]
\centering
\caption{Hyperparameters for the Single-Label Contrastive-Learning based Classification Model}
\label{tab:single_label_classification_hyperparameters}

\begin{tabular*}{\textwidth}{@{\extracolsep{\fill}}lllll}
\toprule
\textbf{Parameter}                & \textbf{PaviaU} & \textbf{Salinas} & \textbf{Houston 2013} & \textbf{Houston 2018} \\ \midrule
\textbf{Epochs}                   & 200             & 200              & 750                   & 400                   \\
\textbf{Batch Size}               & 200             & 164              & 150                   & 250                   \\
\textbf{Learning Rate}            & $1 \times 10^{-3}$  & $2 \times 10^{-3}$  & $1 \times 10^{-3}$  & $1 \times 10^{-3}$      \\
\textbf{Gamma}                    & 0.9             & 0.9              & 0.6                   & 0.9                   \\
\textbf{Lr-Step Size}             & $3 \times 10^{-4}$  & $1 \times 10^{-3}$  & $1 \times 10^{-4}$  & $5 \times 10^{-5}$      \\
\textbf{L2-Regularization Weight} & 10              & 10               & 75                    & 50                    \\
\textbf{Dropout (Classifier)}     & 0.2             & 0.2              & 0.3                   & 0.3                   \\
\textbf{Dropout (Encoder)}        & 0.3             & 0.3              & 0.3                   & 0.3                   \\
\bottomrule
\end{tabular*} 
\end{table*}

 \newpage

\subsection{Data reduction}
\label{sec:data_reduction_results}

In this section, we provide a detailed list of all results achieved across the four datasets, considering two scenarios: 
\begin{itemize}
    \item Different dimensions of the hidden representation. 
    \item Different proportions of available training labelled data.
\end{itemize}

\begin{table}[hbbp!]
\centering
\caption{Multi-label Accuracy for Different Data Proportions and Dimensions}
 
\begin{tabular*}{\textwidth}{@{\extracolsep{\fill}}lllll}
\toprule
\textbf{Dataset-Dimension} & \textbf{100\%} & \textbf{50\%} & \textbf{40\%} & \textbf{20\%} \\ 
\midrule
Pavia (32) & 87.87 & 84.59 & 81.86 & 78.96 \\
Pavia (64) & 88.45 & 85.28 & 84.58 & 80.42 \\
Salinas (32) & 88.86 & 83.06 & 80.77 & 76.44 \\
Salinas (64) & 89.74 & 84.83 & 82.56 & 78.78 \\
Houston 2018 (32) & 85.28 & 83.39 & 83.11 & 81.05 \\
Houston 2018 (64) & 86.46 & 84.51 & 84.03 & 82.28 \\
Houston 2013 (32) & 87.23 & 84.48 & 81.50 & 67.31 \\
Houston 2013 (64) & 87.95 & 85.40 & 84.80 & 77.50 \\
\bottomrule
\end{tabular*} 
\end{table}

\begin{table}[htbp!]
\centering
\caption{Single-label Accuracy for Different Data Proportions and Dimensions}
\begin{tabular*}{\textwidth}{@{\extracolsep{\fill}}lllll}
\toprule
\textbf{Dataset-Dimension} & \textbf{100\%} & \textbf{50\%} & \textbf{40\%} & \textbf{20\%} \\ 
\midrule
Pavia (32) & 94.82 & 90.52 & 90.52 & 84.73 \\
Pavia (64) & 94.17 & 90.05 & 90.77 & 83.94 \\
Salinas (32) & 94.16 & 91.69 & 91.70 & 90.25 \\
Salinas (64) & 93.67 & 92.15 & 92.19 & 89.67 \\
Houston 2018 (32) & 93.47 & 91.38 & 91.26 & 89.50 \\
Houston 2018 (64) & 92.77 & 91.46 & 91.11 & 89.72 \\
Houston 2013 (32) & 91.90 & 89.98 & 88.46 & 83.48 \\
Houston 2013 (64) & 91.37 & 89.88 & 90.51 & 82.64 \\
\bottomrule
\end{tabular*} 
\end{table}

\end{document}